\documentclass[runningheads]{llncs}

\usepackage{eccv}

\usepackage{eccvabbrv}

\usepackage{graphicx}
\usepackage{booktabs}

\usepackage[accsupp]{axessibility}  

\usepackage{hyperref}

\usepackage{orcidlink}

\usepackage{kotex} 
\usepackage{multirow} 
\usepackage[dvipsnames]{xcolor} 
\usepackage{pifont}
\usepackage{bm} 
\usepackage{mathrsfs}   
\usepackage{amssymb}    
\usepackage{algorithm} 
\usepackage{algorithmic} 
\usepackage{booktabs}  
\usepackage{graphicx}  

\usepackage{listings}
\usepackage{float} 
\usepackage{stfloats}
\usepackage{cuted}  
\usepackage{pdflscape}

\newif\ifguideline
\guidelinefalse
\newif\ifcameraready
\camerareadyfalse

\newcommand{\cmark}{\textcolor{green}{\ding{51}}} 
\newcommand{\xmark}{\textcolor{red}{\ding{55}}} 

\colorlet{myblue}{blue} 

\DeclareMathOperator*{\argmax}{arg\,max}
\DeclareMathOperator*{\argmin}{arg\,min}

\begin{document}

\title{Do Flat Minima Improve \\ Sparse Novel View Synthesis?} 


\author{
Youngsik Yun\inst{}\orcidlink{0000-0003-4398-7856}
\and 
Dongjun Gu\inst{}\orcidlink{0000-0001-7804-0254}
\and 
Youngjung Uh\inst{}\thanks{Corresponding author}\orcidlink{0000-0001-8173-3334}
}

\authorrunning{Yun et al.}


\institute{Yonsei University, Seoul, Korea \\
\email{\{bbangsik, djku1020, yj.uh\}@yonsei.ac.kr}
}

\maketitle

\ifguideline
\else

\begin{abstract}
Despite the success of recent novel view synthesis methods, they tend to struggle in sparse-view settings.
This poor generalization to unseen viewpoints is an inherent challenge when training with limited data.
To address this, we investigate the relationship between loss sharpness and generalization in novel view synthesis—an underexplored direction.  
Interestingly, while pursuing flatter minima is widely known to improve generalization in deep learning, reducing loss sharpness is not always  beneficial in novel view synthesis. 
We demonstrate that this difference arises because high-detail regions inherently require a sharp loss landscape for accurate reconstruction, whereas low-detail regions benefit from a flat loss landscape for improving generalization.
Based on this insight, we introduce structure-aware sharpness, defined within structure-adaptive neighborhoods, and propose to adaptively adjust the sharpness regularization weight according to the local image structure.
This strategy encourages flatter minima for generalization while preserving the loss sharpness necessary to reconstruct fine details. 
Across various datasets and configurations, our strategy consistently improves a wide range of baselines.
Code is available at \url{https://bbangsik13.github.io/FASR}.
\end{abstract}

\begin{figure}[t!]
    \centering 
    \includegraphics[width=\linewidth]{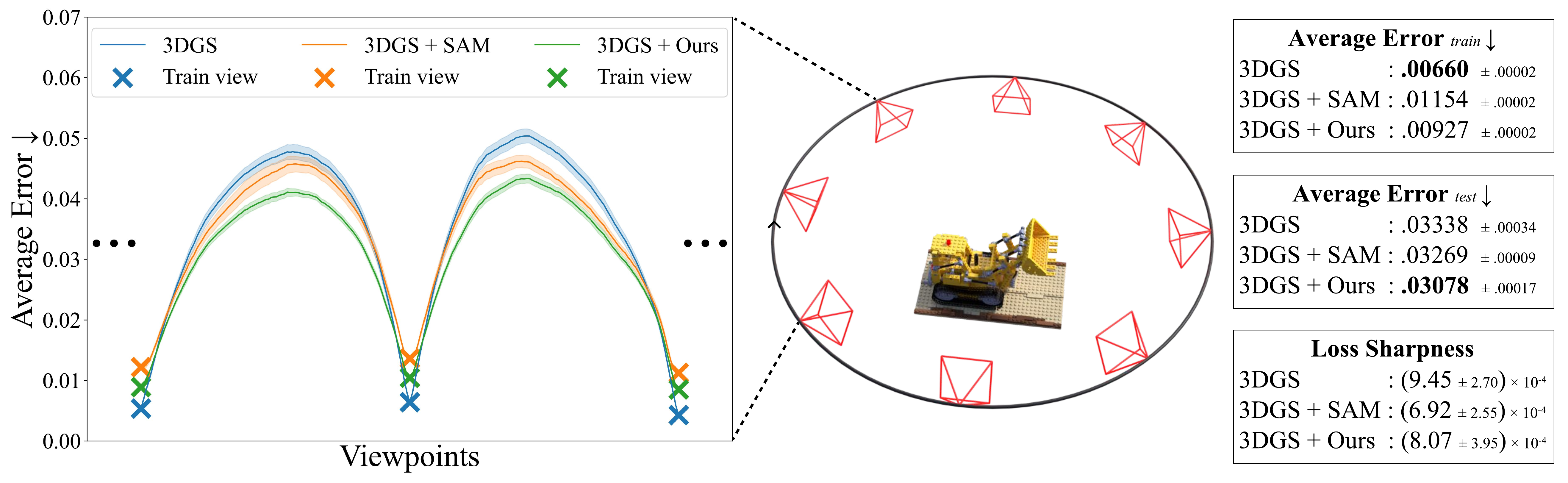} 
    \caption{\textbf{Overview}. While 3DGS exhibits overfitting, our method improves its generalization, achieving lower test errors. Interestingly, loss sharpness and generalization are not strictly correlated in novel view synthesis—although Sharpness-Aware Minimization (SAM)~\cite{foret2021sam}, a representative flat minima optimization method, yields the lowest loss sharpness~\cite{wen2023how,luo2024explicit}, it results in a higher test error than ours—challenging the common belief that flatter minima generalize better. We evaluate using eight training views and 800 interpolated test views generated from the \texttt{lego} scene of Blender synthetic dataset~\cite{mildenhall2020nerf}. Plots are means and standard deviations over ten runs.}
    \label{fig:teaser}
\end{figure}

\section{Introduction}
\label{sec:intro}
Novel view synthesis from multi-view images has been a long-standing problem of interest.
While Neural Radiance Fields (NeRFs)~\cite{mildenhall2020nerf} achieve high-quality novel view synthesis, their slow rendering has motivated a shift towards 3D Gaussian Splatting (3DGS)~\cite{kerbl20233dgs}.
However, it requires densely captured input views, which are costly to obtain.
In sparse-view settings, models often produce unresolved details and floating artifacts in novel viewpoints while correctly reconstructing training views. This overfitting is due to limited data, leading to poor generalization.
To address this challenging problem, previous approaches have adopted various strategies, such as integrating geometric priors~\cite{deng2022dsnerf,dai2025eapgs} and leveraging dense correspondence prediction models~\cite{jang2025comapgs,han2024binocular}. 
Although these methods show effectiveness, research on improving the optimization procedure of such scene representations itself has not been thoroughly explored.

Rather than introducing additional model priors, we investigate the role of local loss sharpness in novel view synthesis. In deep learning, flatter minima\footnote{We refer to flatter \emph{local} minima as flatter minima for brevity.} are widely known to improve generalization~\cite{keskar2017on,izmailov2019swa}.
However, unlike the domain of previous work, achieving flatter minima \emph{does not always guarantee} better generalization in novel view synthesis (\cref{fig:teaser}). 
In \cref{fig:problem}, we demonstrate that this discrepancy arises because the importance of loss sharpness for accurate reconstruction varies across regions: model parameters representing high-detail regions (e.g., edges) inherently induce a sharp loss landscape, whereas those corresponding to low-detail regions favor flatter minima.
In this sense, pursuing flat minima for all regions is problematic; sharp minima in high-detail areas are desirable for accurate reconstruction, and flat minima in low-detail regions are preferable for better generalization. Specifically, it will over-penalize high-detail regions while under-penalize low-detail regions, failing to sufficiently improve generalization.

To this end, we propose Structure-Aware Sharpness Regularization, an optimization strategy that calculates and penalizes loss sharpness based on the local image structure. 
Specifically, we define structure-aware sharpness and adaptively adjust the sharpness regularization: in high-detail regions, we both reduce the regularization weight and the neighborhood radius used for sharpness estimation, while in low-detail regions, we increase them. 
This strategy encourages a flatter loss landscape while \emph{retaining the loss sharpness necessary for fine details}.
For example, our strategy efficiently mitigates floating artifacts that inherently induce sharp minima, because their loss changes dramatically even with a slight perturbation\footnote{Camera perturbations can be interpreted as model parameter perturbations, e.g., parallel movement of 3D Gaussians is equivalent to moving the camera.}. Our strategy guides these unexpected sharp minima to flat minima, successfully narrowing the generalization gap, thus suppressing floaters without penalizing high-detail regions.
Crucially, we demonstrate that our strategy serves as a general principle for novel view synthesis, yielding complementary improvements whether applied to a wide range of 3DGS-based frameworks, implicit NeRF, various flat minima optimization methods, and scene dynamics.

Our core contributions are as follows:
\begin{itemize}
    \item \textbf{Fundamental insight.} To the best of our knowledge, we present the first fundamental investigation into the relationship between loss landscape and generalization in novel view synthesis. Notably, we reveal that flat minima across all Gaussian parameters are suboptimal because the optimal loss sharpness varies with the local structure of the input images.
    \item \textbf{Novel optimization strategy.} Based on this insight, we propose a strategy that introduces structure-aware sharpness and adaptively adjusts the regularization weight to encourage flatness for generalization while preserving the necessary loss sharpness for fine-detail reconstruction.
    \item \textbf{Generality.} We demonstrate that our strategy serves as a broadly applicable optimization principle rather than a method-specific adaptation, consistently improving various reconstruction pipelines.
\end{itemize}

\begin{figure}[tb!]
    \centering
      \includegraphics[width=\linewidth]{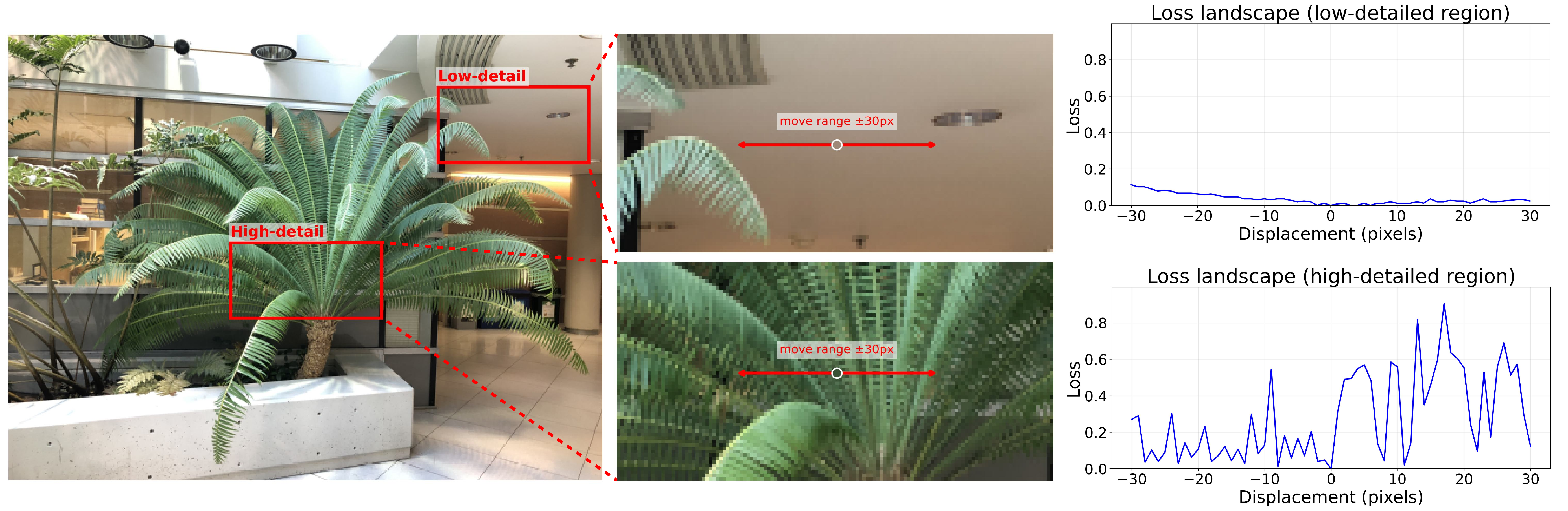}
    \caption{\textbf{Problem setting}. The desirable degree of loss sharpness varies across the local structure of the input images. We visualize the loss landscape by perturbing a perfectly fitted pixel in high-detail and low-detail regions, respectively.}
    \label{fig:problem}
\end{figure}

\section{Related Work}
\label{sec:relwork}
\subsection{Sparse Novel View Synthesis}
\label{sec:relwork:sparse}
Reconstructing scenes from highly sparse inputs remains a fundamental challenge, as limited supervision often leads models to overfit training views, resulting in insufficient generalization in unseen views. 
Early efforts sought to extend neural radiance fields (NeRFs)~\cite{mildenhall2020nerf} by incorporating additional regularizations or auxiliary cues. For example, these studies~\cite{niemeyer2022regnerf,deng2022dsnerf,wang2023sparsenerf,yang2023freenerf} introduce geometry- or depth-based constraints and frequency regularization to alleviate the underconstrained nature of sparse-view optimization. 
Although these approaches demonstrate effectiveness, they inherit the slow training and rendering processes of NeRF.

Recent studies have shifted to use 3D Gaussian Splatting (3DGS)~\cite{kerbl20233dgs}, aiming for real-time rendering efficiency. 
These studies~\cite{li2024dngaussian,zhu2024fsgs,dai2025eapgs,xu2025dropoutgs,jang2025comapgs,zheng2025nexusgs} similarly leverage external priors such as depth~\cite{bhat2023zoedepth,ranftl2021dpt}, correspondence~\cite{leroy2024mast3r}, or flow~\cite{shi2023flowformerpp}, while others focuses on ensemble-like regularization~\cite{zhang2024corgs,park2025dropgaussian,zhao2025segs}.
Sharing a key insight with our method, Sparfels~\cite{jena2025sparfels} aims to minimize a worst-case loss for robustness; however, it does so by freezing the Gaussian means and approximating the objective with an upper bound, which simplifies to a color variance regularization along each ray. Although this proxy effectively improves details, it may overlook color-matched floaters. In contrast, our method directly optimizes the worst-case loss with respect to all Gaussian parameters, interpreting it as a sharpness regularization.

Importantly, our optimization strategy is complementary: rather than altering the representation or adding priors, we fundamentally guide novel view synthesis methods to converge on a general solution in under-constrained sparse supervision.

\subsection{Flatness and Generalization}
\label{sec:relwork:flat}
The relationship between the flatness of the loss landscape and model generalization has been extensively investigated in prior research. Keskar \etal~\cite{keskar2017on} shows that converging to sharp minima leads to poor generalization, while Jiang \etal~\cite{jiang2020fantastic} identifies that loss sharpness is the most correlated indicator of generalization. Subsequent work~\cite{izmailov2019swa,cha2021swad} demonstrated that averaging parameters along training trajectories can lead to flatter minima with better generalization.  

Beyond averaging strategies, Sharpness-Aware Minimization (SAM)~\cite{foret2021sam} explicitly estimates and reduces sharpness, inspiring subsequent work~\cite{andriuschenko2022towards,mueller2023samon,sun2024adasam,li2024fsam} to analyze and improve upon its formulation. 
Moreover, some works have analyzed the accuracy of estimated sharpness as a measure of generalization. They show that sharpness changes with parameter rescalings~\cite{dinh2017sharp}, and the appropriate sharpness estimation differs across different training setups and tasks~\cite{andriushchenko2023modern}, which hinders the correlation between sharpness and generalization.
Consequently, recent work introduces normalization and invariant formulations to appropriately estimate sharpness, enabling a more reliable link between loss sharpness and generalization~\cite{tsuzuku2020normalized,kwon2021asam,kim2022fishersam,jang2022reparametrization,moritz2024up2}.
On the other hand, some works report counterexamples where sharper models generalize well, indicating that flatter minima are not always the optimal strategy for generalization~\cite{dinh2017sharp,wen2023sharpness,andriushchenko2023sharp}.

Building on this perspective, we revisit the loss landscape sharpness in the context of novel view synthesis.
We hypothesize that the optimal loss sharpness is not uniform but varies with the local detail of the input images. Furthermore, the radius for estimating the local loss sharpness should also vary.
Therefore, instead of pursuing flat minima across all model parameters, we introduce an optimization strategy that appropriately regularizes loss sharpness for the reconstruction task, allowing sharp minima where they benefit fine details.

\subsection{Reconstructing with Random Noise}
\label{sec:relwork:perturb}
Adapting random noise into radiance field training has been introduced for two primary purposes. 

Unlike our approach, which aims to improve generalization, some studies employ perturbation for uncertainty quantification rather than relying on a deterministic approach. Stochastic or Bayesian formulations of NeRFs have explicitly modeled radiance or density distributions~\cite{shen2021snerf,lee2025bayesiannerf}. Subsequent work perturbs trained models to estimate epistemic uncertainty~\cite{goli2024bayesrays}.
Similar ideas have been extended to 3DGS, where Gaussian parameters are sampled from learned distributions to render both images and calibrated uncertainty maps~\cite{aira2025stochasticgs}.

Some works~\cite{kheradmand20243dgsmcmc,gao2024hicom,ling2025precondition,chen2025quantifying,seo2026improving} achieve robustness by injecting random noise into Gaussian parameters or query locations, which can be interpreted as randomly choosing parameters in a neighborhood radius. 
This random perturbation serves as an inefficient proxy for finding the worst-case loss, and thus may flatten the loss landscape. 
In contrast, our work fundamentally investigates the direct link between the loss landscape and generalization, allowing us to rethink the success of these prior works.

\begin{algorithm}[tb!]
\caption{Overview of our proposed algorithm}
\label{alg:overview}
\begin{algorithmic}[1]  
  \REQUIRE Multi-view images $\tilde{I}^v$, where camera $v \in \mathcal{V}$ 
  \ENSURE Optimized $\mathcal{G} = (\boldsymbol{\mu}, \mathbf{q}, \mathbf{s}, \sigma, \mathbf{Y}^\text{DC}, \mathbf{Y}^\text{AC})$
  \STATE $\Gamma^v \gets 
  \text{ToleranceMap}(\tilde{I}^v)$\COMMENT{Quantify local image structure (\cref{sec:method:ddmap})}
  \WHILE{$\mathcal{G}$ not converged}
    \STATE $L \gets \text{Loss}(\text{Render}(\mathcal{G}, v), \tilde{I}^v)$ \COMMENT{Empirical loss}
    \FORALL{$\boldsymbol{\theta} \in \mathcal{G}$}
      \STATE $\hat{\boldsymbol{\theta}} \gets \text{StructureAwarePerturb}(\boldsymbol{\theta}, L,\Gamma)$ 
      \COMMENT{Calculate perturbation (\cref{sec:method:fasr:perturb})}
    \ENDFOR
    \STATE $\hat{\mathcal{G}} \gets (\hat{\boldsymbol{\theta}} \mid \boldsymbol{\theta} \in \mathcal{G})$ \COMMENT{Local maximum}
    \STATE $\hat{L} \gets \text{Loss}(\text{Render}(\hat{\mathcal{G}}, v), \tilde{I}^v)$  \COMMENT{Worst-case loss}
    \STATE $\mathcal{L} \gets \text{StructureAwareWeight}(L,\hat{L},\Gamma)$ \COMMENT{Scale regularization weight (\cref{sec:method:fasr:sharp})}
    \STATE $\mathcal{G} \gets \mathcal{G} - \text{Adam}(\nabla_{\hat{\mathcal{G}}}\mathcal{L})$ \COMMENT{Gradient descent at original parameter}
  \ENDWHILE
\end{algorithmic}
\end{algorithm}

\section{Method}
\label{sec:method}
To implement our fundamental insight—optimal loss sharpness for accurate reconstruction correlates with the local structure of the input images—into practice, we apply our strategy within Sharpness-Aware Minimization (SAM)~\cite{foret2021sam} and 3D Gaussian Splatting (3DGS)~\cite{kerbl20233dgs}, as representative choices of flat-minima optimization and scene representation, respectively. First, we provide preliminaries on SAM and 3DGS (\cref{sec:method:prelim}). Then, we introduce the geometric tolerance map to quantify local image structures (\cref{sec:method:ddmap}). Finally, based on this image structure, we define the structure-aware sharpness (\cref{sec:method:fasr:perturb}) and adaptively adjust the regularization weight (\cref{sec:method:fasr:sharp})\footnote{We apply this strategy to the mean $\boldsymbol{\mu}_i$, rotation $\mathbf{q}_i$, and scale $\mathbf{s}_i$, which are geometric attributes of the Gaussian.}. \cref{alg:overview} summarizes our method.

\subsection{Preliminaries}
\label{sec:method:prelim}

\subsubsection{Sharpness-Aware Minimization (SAM)}~\cite{foret2021sam} is an optimization method that improves generalization by encouraging solutions to lie in flat regions of the loss landscape.
Along with minimizing the empirical loss at a parameter point $\boldsymbol{w}$, SAM estimates the loss sharpness within a neighborhood of radius $\rho$. The optimization then jointly minimizes both the empirical loss and this estimated sharpness: 
\begin{equation} 
\begin{split}
L^\text{SAM} \triangleq \underbrace{\max_{\|\boldsymbol{\epsilon}\|_2 \leq \rho} \; L(\boldsymbol{w} + \boldsymbol{\epsilon}) - L(\boldsymbol{w})}_\text{loss sharpness} + \underbrace{L(\boldsymbol{w})}_\text{empirical loss}
= \underbrace{\max_{\|\boldsymbol{\epsilon}\|_2 \leq \rho} \; L(\boldsymbol{w} + \boldsymbol{\epsilon})}_\text{worst-case loss},
\end{split}
\label{eq:sam}
\end{equation} 
which is equivalent to the worst-case loss.

In practice, it is approximated via first-order Taylor expansion for the loss function around the current parameters, perturbing the parameters in the gradient direction with magnitude $\rho$:
\begin{equation}
 \hat{\boldsymbol{\epsilon}}(\boldsymbol{w}) \triangleq \argmax_{\|\boldsymbol{\epsilon}\|_2 \leq \rho} L(\boldsymbol{w}+\boldsymbol{\epsilon}) \approx \rho \cdot \frac{\nabla_{\boldsymbol{w}}L(\boldsymbol{w})}{\|\nabla_{\boldsymbol{w}}L(\boldsymbol{w})\|_2}.
\label{eq:perturbation}
\end{equation}

This method improves generalization by tightening the generalization bound based on sharpness derived from the PAC-Bayesian framework~\cite{taylor1997pacbayesian,mcallester1998pacbayesian,dziugaite2017pacbayesian}.
In practice, $\rho$ is a hyperparameter.
See Foret \etal~\cite{foret2021sam} for details.

\subsubsection{3D Gaussian Splatting (3DGS)}~\cite{kerbl20233dgs} represents a scene as a set of 3D Gaussians $\mathcal{G} = \{\mathcal{G}_i\}_{i=1}^N$. Each Gaussian $\mathcal{G}_i$ is parameterized by its mean $\boldsymbol{\mu}_i$, rotation $\mathbf{q}_i$, scale $\mathbf{s}_i$, opacity $\sigma_i$, and spherical harmonics (SH) coefficients $\mathbf{Y}_i^{\mathrm{DC}}, \mathbf{Y}_i^{\mathrm{AC}}$ that model view-dependent color.
Given a camera view $v$, the renderer projects each Gaussian onto the image plane and produces a rendered image $I^v = \mathrm{Render}(\mathcal{G}, v)$.
See Kerbl \etal~\cite{kerbl20233dgs} for details. 

\subsection{Geometric Tolerance Map}
\label{sec:method:ddmap}
To quantify the local structure of the input images, we introduce a geometric tolerance map $\Gamma^v(\mathbf{p})$ defined over each training image $\tilde{I}^v$ with view $v$, where $\mathbf{p}$ denotes a 2D image coordinate.
The map encodes the distance from each pixel to the nearest high-detail structure, such as fine-scale textures.
Let $\mathcal{H}^v$ denote the set of pixels with large image gradient in view $v$.
Then, the map is defined as:
\[
\Gamma^v(\mathbf{p}) = \min_{\mathbf{h} \in \mathcal{H}^v} \|\mathbf{p} - \mathbf{h}\|_2.
\]
As a result, pixels located in low-detail structure exhibit large distance values, whereas pixels near or within high-detail structure have small values.

Near high-detail structure, even small geometric changes of projected primitives can produce noticeable variations in the rendered image.
Conversely, in low-detail structure, similar changes tend to have a limited visual impact.
Therefore, we interpret this distance as a \emph{geometric tolerance} for the primitive projected at location $\mathbf{p}$, with respect to rendering sensitivity.

We describe the implementation of the geometric tolerance map in the Appendix.

\subsection{Structure-Aware Sharpness}
\label{sec:method:fasr:perturb}
We adapt the loss sharpness of each Gaussian based on the geometric tolerance defined in \cref{sec:method:ddmap}.
To do this, we begin by computing the gradient on the per-Gaussian attribute $\boldsymbol{\theta}_i \in \mathcal{G}_i$ separately rather than the entire parameter set and use it to compute the perturbation independently (\cref{eq:perturbation}).

Then, for each Gaussian, we query the value from the geometric tolerance map at the projected center coordinate $\mu'_i$ on the 2D camera plane, $\boldsymbol{\gamma}_i = \Gamma^v(\mu'_i)$.
Since this geometric tolerance is defined in the 2D image space, we convert it into a 3D space using perspective geometry.
Under perspective projection, a 2D displacement of $\boldsymbol{\gamma}_i$ corresponds to a 3D displacement scaled by $\frac{d_i}{f}$, where $d_i$ is the depth of the corresponding Gaussian and $f$ is the focal length.
We interpret this 3D geometric tolerance as an admissible perturbation of the 3D parameters of the Gaussian.
Accordingly, the structure-aware sharpness for the Gaussian attribute $\boldsymbol{\theta}_i$ is defined as:
\[
\max_{\|\boldsymbol{\epsilon}_i\|_2 \le \boldsymbol{\gamma}_i \frac{d_i}{f}\rho_{\boldsymbol{\theta}}}
L(\boldsymbol{\theta}_i + \boldsymbol{\epsilon}_i)
-
L(\boldsymbol{\theta}_i).
\]

In the SAM implementation, the perturbation \cref{alg:overview}.5 becomes:
$$
\hat{\boldsymbol{\theta}}_i \gets \boldsymbol{\theta}_i + \boldsymbol{\gamma}_i \frac{d_i}{f}\rho_{\boldsymbol{\theta}} \dfrac{\nabla_{\boldsymbol{\theta}_i} L}{\|\nabla_{\boldsymbol{\theta}_i} L\|_2}.
\label{eq:mu_radius}
$$
This structure-aware formulation resolves the imbalance in the perturbation magnitude. Specifically, it induces slight perturbations in high-detail regions to prevent overshoot of the intended local maximum, while applying stronger perturbations in low-detail regions to effectively ascend toward the local maximum.

We provide a deeper analysis of this derivation in the Appendix.

\subsection{Structure-Aware Regularization Weighting}
\label{sec:method:fasr:sharp}
In addition to structure-aware sharpness, we also modulate the weight of sharpness regularization.
As described in \cref{eq:sam}, SAM can be interpreted as minimizing empirical loss together with a sharpness term.
To control the relative contribution of these two terms, we introduce a structure-aware weighting scheme.

Inspired by weighted variants of SAM~\cite{yue2023wsam}, we consider a weighted objective of the following form. Therefore, \cref{alg:overview}.9 becomes:
\[
\mathcal{L} = L + \frac{\bar{\boldsymbol{\gamma}}_i}{1-\bar{\boldsymbol{\gamma}}_i} \, (\hat{L}-L) = \frac{1-2\bar{\boldsymbol{\gamma}}_i}{1-\bar{\boldsymbol{\gamma}}_i} L + \frac{\bar{\boldsymbol{\gamma}}_i}{1-\bar{\boldsymbol{\gamma}}_i}\hat{L}.
\]
Let the regularization weight $\bar{\boldsymbol{\gamma}}_i = 0.95 \boldsymbol{\gamma}_i / \boldsymbol{\gamma}_\text{max}$, where $\boldsymbol{\gamma}_\text{max}$ is the maximum distance, and 0.95 is determined empirically.
When the regularization weight $\bar{\boldsymbol{\gamma}}i$ is close to 0, the sharpness term vanishes, and the objective reduces to minimizing only the empirical loss. In contrast, when it is close to 0.95, the sharpness term is emphasized.

As a result, the sharpness penalty is reduced in high-detail regions, preserving necessary loss sharpness for accurate reconstruction, while applying a stronger penalty in low-detail regions to facilitate better regularization.

\begin{figure*}[tb!]
    \centering
    \includegraphics[width=\textwidth]{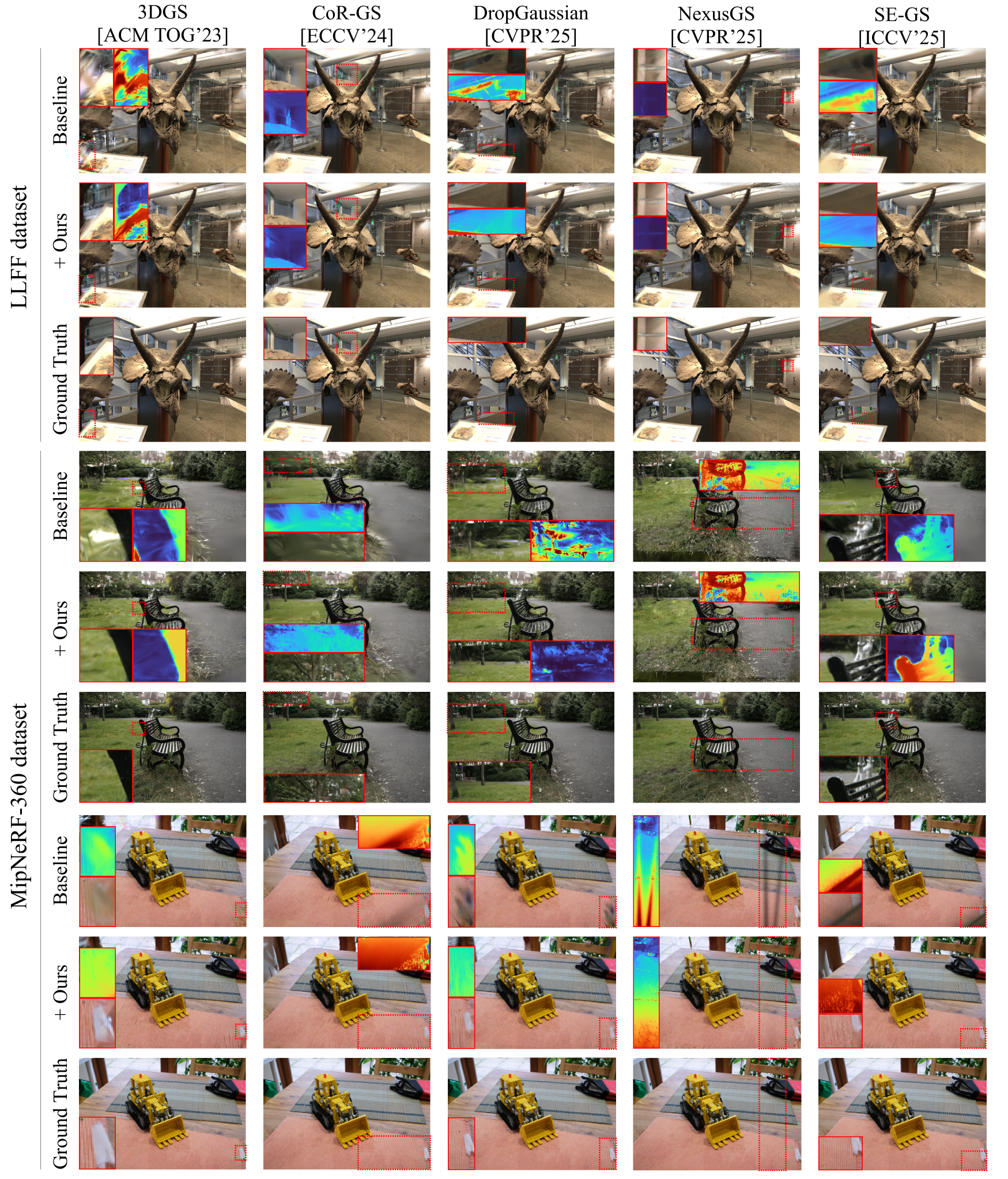}
    \caption{\textbf{Qualitative comparison}.
    Our method can be applied to various baselines in a plug-and-play manner, improving reconstruction quality by removing floaters while preserving fine details.}
    \label{fig:qual}
\end{figure*}

\begin{table}[tb!]
    \centering
    \caption{\textbf{Quantitative comparison}. Our method consistently improves all baselines across both LLFF and Mip-NeRF360, achieving better results on all evaluation metrics.} 
    \resizebox{\linewidth}{!}{
    \begin{tabular}{lcccccc}
    \toprule
     & \multicolumn{3}{c}{LLFF (3 views)} & \multicolumn{3}{c}{MipNeRF-360 (12 views)}  \\ \cmidrule(lr){2-4}\cmidrule(lr){5-7}
    Method & PSNR~$\uparrow$ & SSIM~$\uparrow$ & LPIPS~$\downarrow$ & PSNR~$\uparrow$ & SSIM~$\uparrow$ & LPIPS~$\downarrow$ \\
    \midrule
    3DGS [ACM TOG'23] & 19.810 \scriptsize{$\pm$ .339} & .6790 \scriptsize{$\pm$ .0078} & .2145 \scriptsize{$\pm$ .0065}  & 18.903 \scriptsize{$\pm$ .179} & .5499 \scriptsize{$\pm$ .0036} & .3734 \scriptsize{$\pm$ .0042} \\
    + Ours & \textbf{20.783} \scriptsize{$\pm$ .300} & \textbf{.7197} \scriptsize{$\pm$ .0032} & \textbf{.1965} \scriptsize{$\pm$ .0034}  &  \textbf{19.303}  \scriptsize{$\pm$ .185} & \textbf{.5622}  \scriptsize{$\pm$ .0051} & \textbf{.3552}  \scriptsize{$\pm$ .0047}  \\
    \midrule
    CoR-GS [ECCV'24] & 20.185 \scriptsize{$\pm$ .142} & .7015 \scriptsize{$\pm$ .0040} & .2029 \scriptsize{$\pm$ .0035} & 19.515 \scriptsize{$\pm$ .243} & .5733 \scriptsize{$\pm$ .0049} & .3741 \scriptsize{$\pm$ .0066}  \\
    + Ours & \textbf{20.862} \scriptsize{$\pm$ .154} & \textbf{.7283} \scriptsize{$\pm$ .0042} & \textbf{.1932} \scriptsize{$\pm$ .0030}   &  \textbf{19.805}  \scriptsize{$\pm$ .233} & \textbf{.5833}  \scriptsize{$\pm$ .0058} & \textbf{.3681}  \scriptsize{$\pm$ .0069}   \\
    \midrule
    DropGaussian [CVPR'25] & 20.461 \scriptsize{$\pm$ .212} & .7070 \scriptsize{$\pm$ .0045} & .2064 \scriptsize{$\pm$ .0047} & 19.514 \scriptsize{$\pm$ .199} & .5722 \scriptsize{$\pm$ .0042} & .3657 \scriptsize{$\pm$ .0036}  \\
    + Ours & \textbf{20.853} \scriptsize{$\pm$ .227} & \textbf{.7295} \scriptsize{$\pm$ .0045} & \textbf{.1969} \scriptsize{$\pm$ .0040}   & \textbf{19.625}  \scriptsize{$\pm$ .254} & \textbf{.5750}  \scriptsize{$\pm$ .0053} & \textbf{.3627}  \scriptsize{$\pm$ .0051}   \\
    \midrule
    NexusGS [CVPR'25] & 21.048 \scriptsize{$\pm$ .049} & .7382 \scriptsize{$\pm$ .0008} & .1776 \scriptsize{$\pm$ .0009} & 18.506 \scriptsize{$\pm$ .098} & .5222 \scriptsize{$\pm$ .0031} & .3587 \scriptsize{$\pm$ .0021} \\
    + Ours & \textbf{21.348} \scriptsize{$\pm$ .078} & \textbf{.7511} \scriptsize{$\pm$ .0012} & \textbf{.1714} \scriptsize{$\pm$ .0011}   &  \textbf{18.736}  \scriptsize{$\pm$ .103} & \textbf{.5316}  \scriptsize{$\pm$ .0034} & \textbf{.3522}  \scriptsize{$\pm$ .0024}  \\
    \midrule
    SE-GS [ICCV'25] & 20.725 \scriptsize{$\pm$ .217} & .7203 \scriptsize{$\pm$ .0049} & .1861 \scriptsize{$\pm$ .0058} & 19.931 \scriptsize{$\pm$ .288} & .5930 \scriptsize{$\pm$ .0063} & .3702 \scriptsize{$\pm$ .0054} \\
    + Ours & \textbf{21.141} \scriptsize{$\pm$ .223} & \textbf{.7403} \scriptsize{$\pm$ .0042} & \textbf{.1803} \scriptsize{$\pm$ .0038}   &  \textbf{20.135}  \scriptsize{$\pm$ .232} & \textbf{.5960}  \scriptsize{$\pm$ .0059} & \textbf{.3644}  \scriptsize{$\pm$ .0052}  \\
    \bottomrule
    \end{tabular}
    }
    \label{tab:static}
\end{table}

\section{Experiments}
\label{sec:exp}

\paragraph{Dataset.}
\label{sec:exp:setting:dataset}
We evaluate our method on LLFF~\cite{ben2019llff} and MipNeRF-360~\cite{barron2022mipnerf360}, following previous works~\cite{zhu2024fsgs,zhang2024corgs,park2025dropgaussian,zheng2025nexusgs}, where the input resolution is 8$\times$ downsampled, and 3 and 12 input views are split for LLFF and MipNeRF-360, respectively.

\paragraph{Implementation.}
\label{sec:exp:setting:implementation}
We choose the publicly available 3DGS~\cite{kerbl20233dgs} and its follow-up works as baselines. Specifically, we choose the state-of-the-art NexusGS~\cite{zheng2025nexusgs}, which leverages foundation models, and CoR-GS~\cite{zhang2024corgs}, DropGaussian~\cite{park2025dropgaussian}, and SE-GS~\cite{zhao2025segs}, which do not.

\paragraph{Metrics.}
\label{sec:exp:setting:metric}
We use PSNR, SSIM~\cite{wang2004ssim}, and LPIPS~\cite{zhang2018lpips} as evaluation metrics, where LPIPS is computed with a VGG network~\cite{simonyan2015vgg}. Additionally, we use Average Error (AVGE)~\cite{niemeyer2022regnerf}, which is the geometric mean of PSNR, SSIM, and LPIPS. 
Considering the randomness of 3DGS, we conduct ten runs on the LLFF dataset and five runs on the MipNeRF-360 dataset.
We provide an analysis of computational cost in the Appendix.

\subsection{Reconstruction Quality}
\label{sec:exp:performance}
As shown in \cref{tab:static}, our method achieves clear gains
in all metrics, datasets, and baselines.
Importantly, these improvements are achieved without any architectural changes or additional priors, but only with our optimization algorithm.
\cref{fig:qual} shows visual comparisons. Applying our method consistently improves the baselines by correcting geometric inaccuracies and reducing floating artifacts in novel viewpoints.
These results reveal that our proposed optimization algorithm can be seamlessly integrated into current and future 3DGS-based frameworks, providing complementary enhancements.

\begin{figure*}[tb!]
    \centering
    \includegraphics[width=0.8\textwidth]{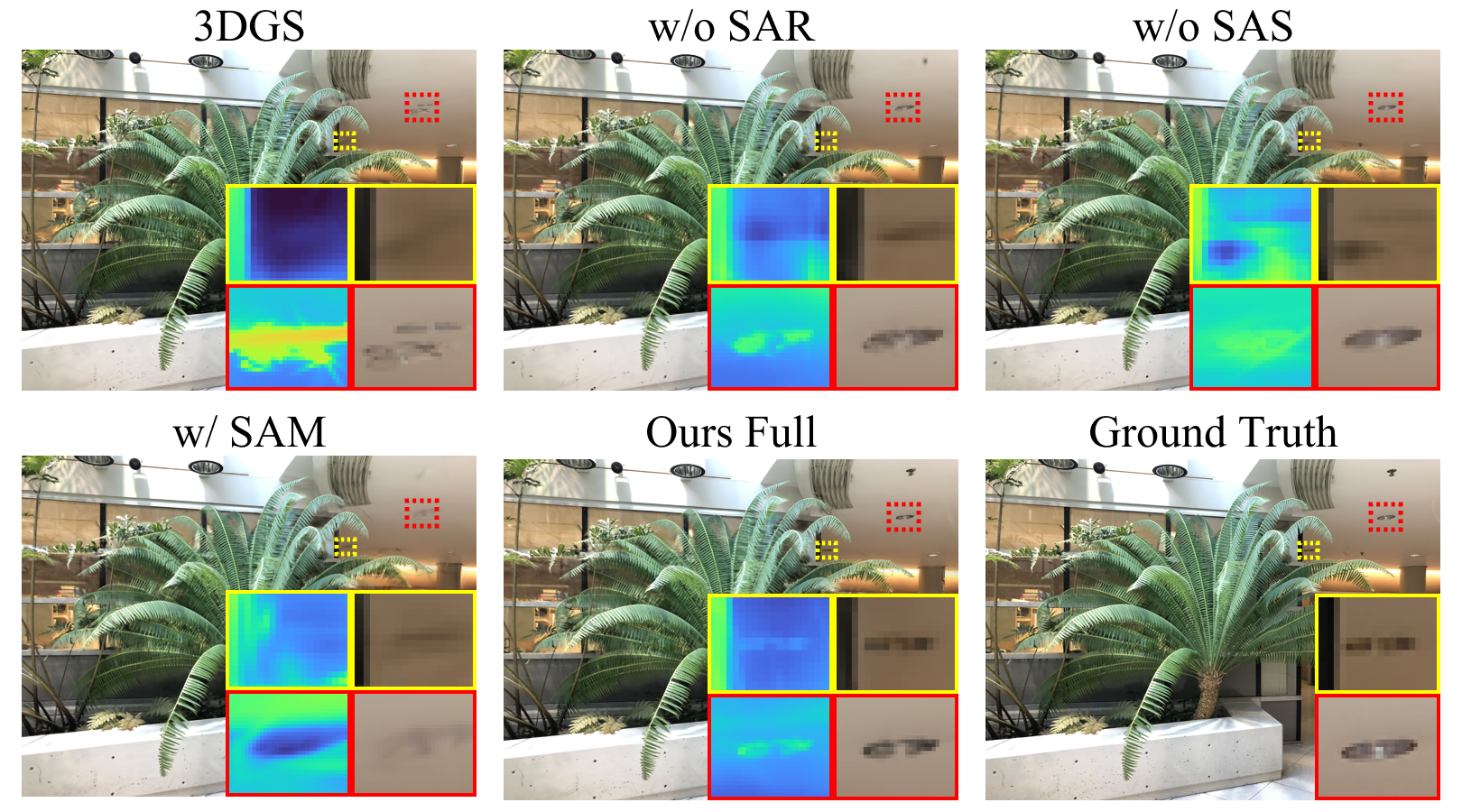}
    \caption{\textbf{Ablation study on LLFF dataset}. SAR and SAS denote structure-aware regularization weight and structure-aware sharpness, respectively. ``3DGS'' and ``SAM'' produce inaccurate geometry (red box). All except ``Ours Full'' show blurry results (yellow box).}
    \label{fig:qual_ablation}
\end{figure*}
\begin{table}[tb!]
    \centering
    \caption{\textbf{Ablation study on LLFF dataset}. SAS and SAR denote structure-aware sharpness and structure-aware regularization weight, respectively.}
    \begin{tabular}{cccccc}
    \toprule
    SAM & SAS & SAR & PSNR~$\uparrow$ & SSIM~$\uparrow$ & LPIPS~$\downarrow$ \\
    \midrule
    \xmark & \xmark & \xmark & 19.810 \scriptsize{$\pm$.339} & .6790 \scriptsize{$\pm$.0078}  & .2145 \scriptsize{$\pm$.0065} \\
    \midrule
    \cmark & \xmark & \xmark & 20.198 \scriptsize{$\pm$.218}  & .6958 \scriptsize{$\pm$.0034} & .2095 \scriptsize{$\pm$.0036} \\
    \midrule
    \cmark & \xmark & \cmark & 20.570 \scriptsize{$\pm$.161} & .6980 \scriptsize{$\pm$.0037} & .2142 \scriptsize{$\pm$.0042}  \\
    \cmark & \cmark & \xmark & 20.560 \scriptsize{$\pm$.259} & .7116 \scriptsize{$\pm$.0038} & .2023 \scriptsize{$\pm$.0032}   \\
    \midrule
    \cmark & \cmark & \cmark &  \textbf{20.783} \scriptsize{$\pm$.300} & \textbf{.7197} \scriptsize{$\pm$.0032} & \textbf{.1965} \scriptsize{$\pm$.0034} \\
    \bottomrule
    \end{tabular}
    \label{tab:ablation}
\end{table}

\subsection{Analysis}
\label{sec:exp:analysis}

\subsubsection{Ablation study.}
\label{sec:exp:analysis:ablation}
As shown in \cref{fig:qual_ablation} and \cref{tab:ablation}, applying SAM~\cite{foret2021sam} directly to 3DGS leads to degraded performance. This naive approach uniformly perturbs Gaussians regardless of image structure, resulting in blurry reconstructions.
Structure-aware regularization weighting preserves sharpness in high-detail regions. Meanwhile, structure-aware sharpness enables more faithful estimation by adaptively adjusting perturbation magnitude. Applying either component individually shows performance gains.
However, removing either component degrades performance. Using both achieves better generalization, balancing between sharpness reduction and detail preservation.

\subsubsection{Loss landscape visualization.}
\label{sec:exp:analysis:landscape}
To analyze the convergence behavior of our method, we visualize the reconstruction loss landscape and the corresponding optimization trajectories.

SAM converges to flatter minima than 3DGS. In \cref{fig:loss_landscape}a, the loss range between the local maximum and minimum is $2.37\times$, and the measured loss sharpness $\lambda_\text{max}$ is $1.49\times$ smaller. SAM also achieves a test loss of 0.0129 lower than 3DGS, narrowing the generalization gap from 0.0985 to 0.0809. 
However, our method converges to less flat minima than SAM. As shown in \cref{fig:loss_landscape}b, the local loss range and measured loss sharpness of ours are $1.11\times$ smaller and $1.01\times$ smaller than 3DGS, respectively. Interestingly, our method achieves test loss of 0.0141 lower than 3DGS and further narrows the generalization gap to 0.0805, outperforming SAM in generalization. 

These results suggest that loss sharpness is not strictly correlated with generalization, supporting our hypothesis that loss sharpness of high-detail regions should be preserved. This aligns with our finding that SAM tends to over-penalize fine details, leading to blurry results (\cref{fig:qual_ablation}).

\begin{figure}[tb!]
    \centering
    \includegraphics[width=0.85\linewidth]{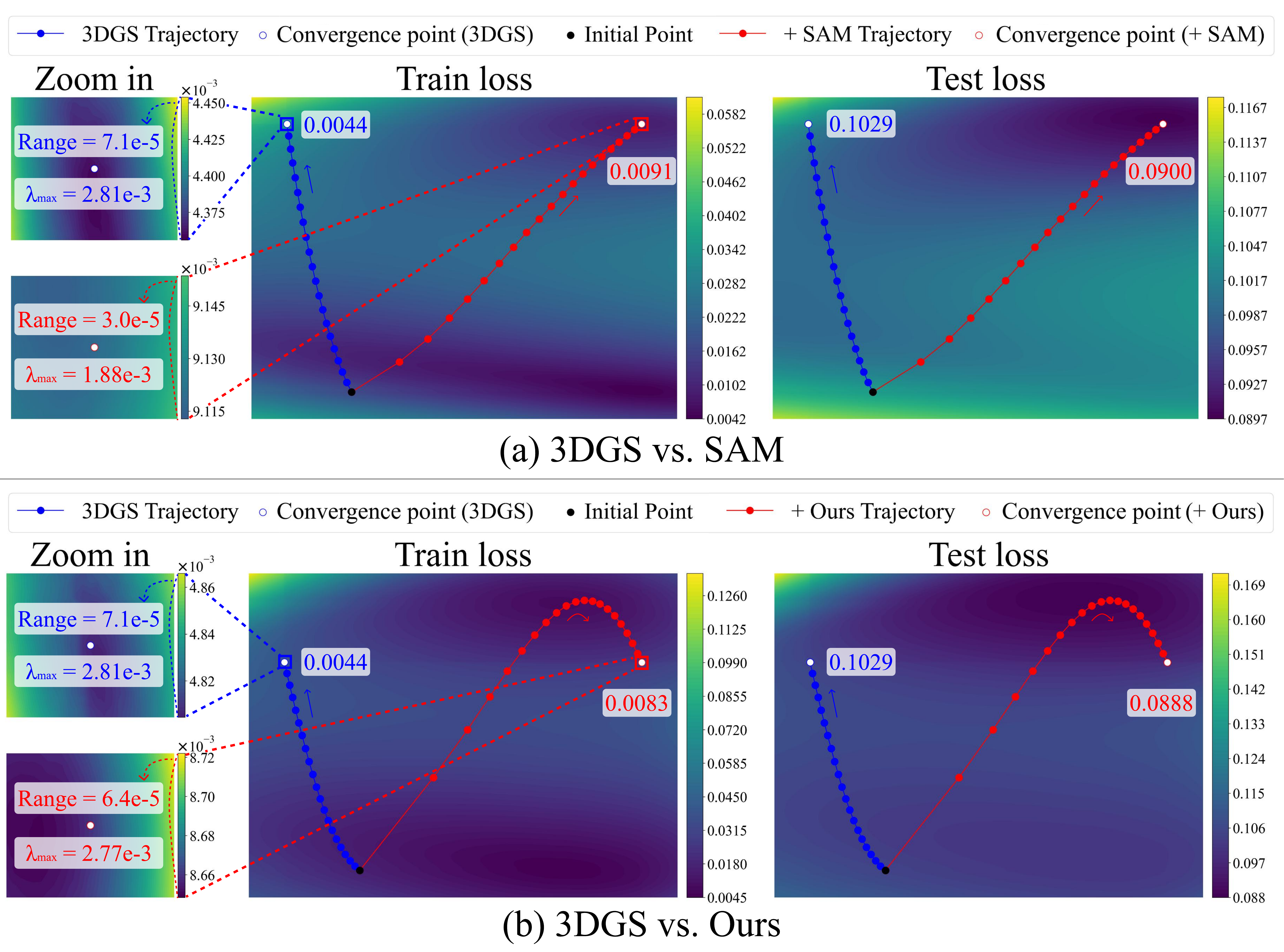}
    \caption{\textbf{Loss landscape visualization on the LLFF \texttt{fern} scene}. We compare the convergence behaviors of 3DGS, SAM, and Ours. 
    To do this, we train 3DGS for 5k iterations, and continue training for an additional 5k iterations without densification, under three distinct settings: 3DGS optimization, SAM, and our proposed method. For visualization, we project the high-dimensional parameter trajectories onto a 2D plane using Principal Component Analysis with parameters from both trajectories. Because the visualization produces a smoothed loss landscape, we provide a zoomed-in view near the convergence points. We measure loss sharpness as the maximum eigenvalue $\lambda_\text{max}$ of the Hessian matrix~\cite{wen2023how,luo2024explicit}.
    }
    \label{fig:loss_landscape}
\end{figure}

\begin{table}[tb!]
    \centering
    \caption{\textbf{Performance by covisibility level on LLFF dataset}. Our method shows greater improvement with higher view sparsity.}
    \begin{tabular}{lccc}
    \toprule
    Covisibility level & 3DGS & \multicolumn{1}{c}{+ Ours} & $\Delta$ \\
    \midrule
Covisibility 3 & .0483 \scriptsize{$\pm$.0093} & .0417 \scriptsize{$\pm$.0080} & -.0066 \scriptsize{$\pm$.0043} \\
Covisibility 2 & .0757 \scriptsize{$\pm$.0156} & .0644 \scriptsize{$\pm$.0137} & -.0114 \scriptsize{$\pm$.0067} \\
Covisibility 1 & .0948 \scriptsize{$\pm$.0256} & .0806 \scriptsize{$\pm$.0232} & \textbf{-.0142} \scriptsize{$\pm$.0074} \\
    \bottomrule
    \end{tabular}
    \label{tab:covis}
\end{table}

\subsubsection{Performance improvement by covisibility level.}
\label{sec:exp:analysis:covis}
The improvement from our method is more substantial in regions observed by fewer training views.
Following CoMapGS~\cite{jang2025comapgs}, we compute covisibility maps using MASt3R~\cite{leroy2024mast3r}.
As shown in  \cref{tab:covis}, the improvement of our method over the baseline 3DGS increases progressively as the covisibility decreases.
This behavior aligns with our hypothesis that our method enhances generalization, particularly in under-constrained regions.

\subsection{Generality of Our Insight}
\label{sec:exp:app}
While our primary evaluation focuses on SAM and 3DGS, our core insight—the optimal loss sharpness varies with local detail—has broader implications for reconstruction tasks. Below, we demonstrate its generality with another flat minima optimizer, dynamic scenes, and the NeRFs representation.

\subsubsection{Compatibility with another flat minima optimization method.}
To validate the broader applicability, we apply our core insight to another flat minima optimization technique, Random Weight Perturbation (RWP)~\cite{li2024revisiting} instead of SAM. Specifically, within the mixed-RWP loss objective~\cite{li2024revisiting}, we set both perturbation magnitude and the balance coefficient to be adaptive to local image structure. 

As shown in \cref{tab:rwp}, integrating our approach into RWP yields consistent performance improvements, echoing the quantitative gains in \cref{tab:ablation} (third and sixth rows). Furthermore, \cref{fig:rwp} illustrates that applying our approach to RWP effectively improves fine details, mirroring the results presented in \cref{fig:qual_ablation} (second row).
These results align with our main experiments, further suggesting that flat minima optimization techniques become effective for reconstruction tasks when applied with our insight.

\begin{table}[tb!]
    \centering
    \caption{\textbf{Quantitative comparison}. Our approach generalizes well to other flat minima optimization method.}
    \resizebox{\linewidth}{!}{
    \begin{tabular}{lcccccc}
    \toprule
     & \multicolumn{3}{c}{LLFF (3 views)} & \multicolumn{3}{c}{MipNeRF-360 (12 views)}  \\ \cmidrule(lr){2-4}\cmidrule(lr){5-7}
    Method & PSNR~$\uparrow$ & SSIM~$\uparrow$ & LPIPS~$\downarrow$ & PSNR~$\uparrow$ & SSIM~$\uparrow$ & LPIPS~$\downarrow$ \\
    \midrule
    3DGS + RWP & 20.079 \scriptsize{$\pm$ .278} & .6971 \scriptsize{$\pm$ .0050} & .1984 \scriptsize{$\pm$ .0039}  & 19.090 \scriptsize{$\pm$ .187} & .5544 \scriptsize{$\pm$ .0047} & .3767 \scriptsize{$\pm$ .0042} \\
    + Ours & \textbf{20.650} \scriptsize{$\pm$ .280} & \textbf{.7157} \scriptsize{$\pm$ .0043} & \textbf{.1912} \scriptsize{$\pm$ .0039}  &  \textbf{19.284}  \scriptsize{$\pm$ .193} & \textbf{.5664}  \scriptsize{$\pm$ .0043} & \textbf{.3567}  \scriptsize{$\pm$ .0034}  \\
    \bottomrule
    \end{tabular}
    }
    \label{tab:rwp}
\end{table}

\begin{figure}[t!]
    \centering
    \includegraphics[width=0.8\linewidth]{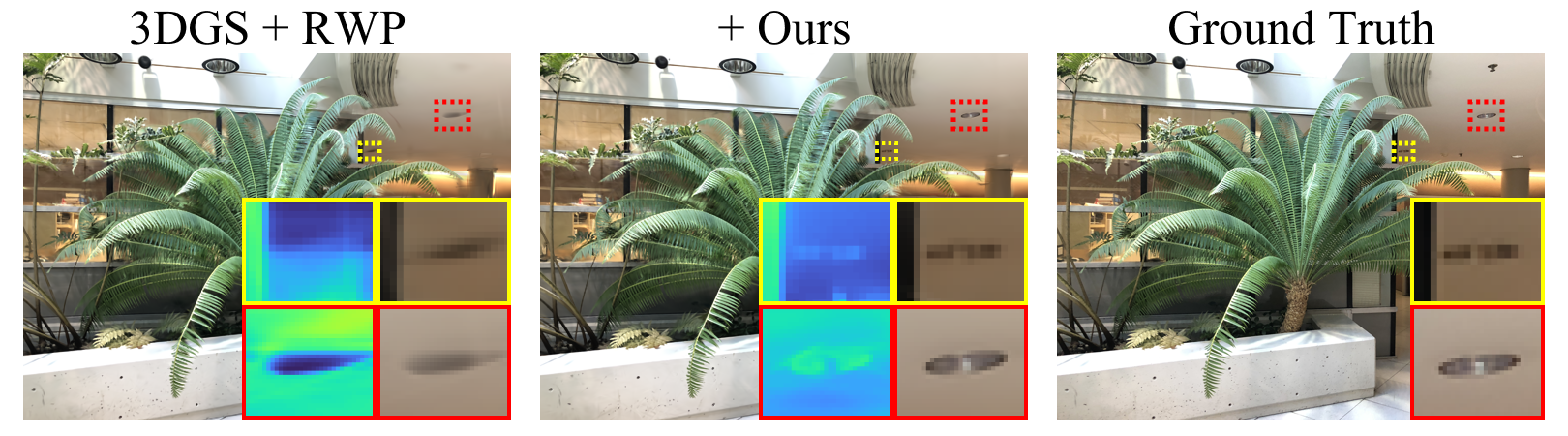}
    \caption{\textbf{Qualitative results demonstrating the effect of our approach on another flat minima optimization method}. RWP itself produces blurry results, whereas applying our approach improves fine details.}
    \label{fig:rwp}
\end{figure}

\begin{table}[tb!]
    \centering
    \caption{\textbf{Applying FASR to online dynamic 3D Gaussians}. Our method is effective in dynamic scenarios under temporal sparsity, notably improving temporal consistency by reducing mTV.}
    \begin{tabular}{lccc}
    \toprule
    Method & PSNR~$\uparrow$ & SSIM~$\uparrow$ & mTV~$\downarrow$ \\
    \midrule
    Yun \etal~\cite{yun2025or2} & 32.542 & .9486  & .1109  \\
    + Ours & \textbf{32.622} & \textbf{.9497} & \textbf{.0989}  \\
    \bottomrule
    \end{tabular}
    \label{tab:online}
\end{table}

\subsubsection{Improving generalization in temporal sparsity.}
\label{sec:exp:app:online}
We demonstrate the extensibility of our insight to dynamic scenes. Specifically, we conduct experiments on the Neural 3D Video dataset~\cite{li2022n3v} with an online configuration~\cite{li2022streamrf,girsh2024queen,hu20254dgc,yan2025igs}, where observations are spatially dense but temporally sparse, meaning that we can only access the current frame in a sequentially processed video stream.
We applied our method to Yun \etal~\cite{yun2025or2} with the 3DGStream~\cite{sun20243dgstream} backbone. 
Following their protocol, we select the first frame 3D Gaussians with the highest PSNR for initialization and compute the masked total variation (mTV) to measure temporal consistency.

Our method improves temporal consistency and visual quality compared to the baseline (\cref{tab:online}). This shows that our approach enhances generalization not only in the spatial domain but also in the temporal domain. 
Furthermore, Yun \etal~\cite{yun2025or2} claims that one cause of temporal jittering is the inevitable noise in training datasets. Since Foret \etal~\cite{foret2021sam} demonstrate robustness in noisy training data, our finding aligns with this explanation.

\subsubsection{Applying our insight to NeRFs.}
\label{sec:exp:app:nerf}
To validate applicability, we apply our core insight to another scene representation, NeRFs~\cite{mildenhall2020nerf}.
As NeRFs are an implicit representation where each parameter influences the entire scene, we apply our insight to baselines that employ a coarse-to-fine procedure~\cite{yang2023freenerf,ling2025precondition}.
Specifically, we set a strong perturbation magnitude and regularization weight when the model parameters learn low-frequency components, and gradually decrease these values as the model learns high-frequency details.

We apply this approach to FreeNeRF~\cite{yang2023freenerf} and demonstrate that it improves both visual quality in novel view synthesis (\cref{fig:nerf}) and quantitative results (\cref{tab:nerf}). Although the improvement over the NeRF baseline is smaller than that in the 3DGS baselines due to inherent differences in the representation, these additional gains suggest that our core insight remains effective regardless of the specific implementation or underlying scene representation.

\begin{table}[t]
        \centering
        \caption{\textbf{Quantitative comparison with FreeNeRF on LLFF dataset}. Applying our approach to the NeRF baseline provides additional performance gain.}
        \begin{tabular}{lccc}
        \toprule
        Method & PSNR~$\uparrow$ & SSIM~$\uparrow$ & LPIPS~$\downarrow$ \\
        \midrule
        FreeNeRF & 19.523 & .6063  & .3103   \\
        + Ours & \textbf{19.584} & \textbf{.6182} & \textbf{.2983} \\
        \bottomrule
        \end{tabular}
        \label{tab:nerf}
\end{table}
\begin{figure}[t]
    \centering
    \includegraphics[width=0.8\linewidth]{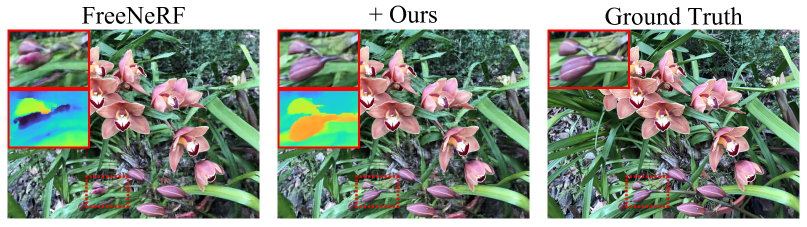}
        \caption{\textbf{Qualitative comparison with FreeNeRF on LLFF dataset}. Applying our approach corrects the inaccurate geometry of FreeNeRF.
        }
        \label{fig:nerf}
\end{figure}

\section{Conclusion}
\label{sec:conclusion}
In this work, we present the first fundamental investigation linking loss landscape and generalization to sparse novel view synthesis.
Notably, we provide valuable insight that optimal loss sharpness in novel view synthesis is highly correlated with local image structure: both flat and sharp minima are essential for accurate reconstruction, which challenges the typical perspective that flatter minima generalize better.
To implement this insight, we introduce Structure-Aware Sharpness Regularization, an optimization algorithm that adapts both the neighborhood radius when calculating loss sharpness and the regularization weight, considering the local image structure. 
Our approach improves baseline performance by correcting geometric inaccuracies and suppressing floating artifacts. 
We demonstrate these consistent improvements across a wide range of 3DGS-based frameworks, various flat minima optimization methods, implicit and explicit representations, and static and dynamic scenarios.
These results suggest that our strategy provides a general and practical principle for novel view synthesis rather than a narrow implementation.
We hope this work inspires the research community to explore the link between loss sharpness and generalization in novel view synthesis, opening a new research direction. 

\section*{Acknowledgements}
This work was supported by an Institute for Information \& communications Technology Planning \& Evaluation (IITP) grant funded by the Korea government (MSIT) (No. RS-2017-II170072, Development of Audio/Video Coding and Light Field Media Fundamental Technologies for Ultra Realistic
Tera-media)

\bibliographystyle{splncs04}
\bibliography{main}

\ifcameraready
\nocite{jiang2025anysplat}
\nocite{jensen2014dtu}
\nocite{zhou2025latesam}
\nocite{lindeberg2013log}
\nocite{lu2023ommo}
\else
\clearpage
\appendix
\setcounter{page}{1}

\renewcommand{\thetable}{A\arabic{table}}
\renewcommand{\thefigure}{A\arabic{figure}}
\renewcommand{\theequation}{A\arabic{equation}}
\setcounter{figure}{0}
\setcounter{table}{0}
\setcounter{equation}{0}
\renewcommand{\theHtable}{A\arabic{table}}
\renewcommand{\theHfigure}{A\arabic{figure}} 
\renewcommand{\theHequation}{A\arabic{equation}} 

\title{Supplementary for Do Flat Minima Improve Sparse Novel View Synthesis?} 

\author{
Youngsik Yun\inst{}\orcidlink{0000-0003-4398-7856}
\and 
Dongjun Gu\inst{}\orcidlink{0000-0001-7804-0254}
\and 
Youngjung Uh\inst{}\thanks{Corresponding author}\orcidlink{0000-0001-8173-3334}
}

\authorrunning{Yun et al.}


\institute{Yonsei University, Seoul, Korea \\
\email{\{bbangsik, djku1020, yj.uh\}@yonsei.ac.kr}
}

\maketitle

\begin{figure}[h!]
    \centering
    \includegraphics[width=0.6\linewidth]{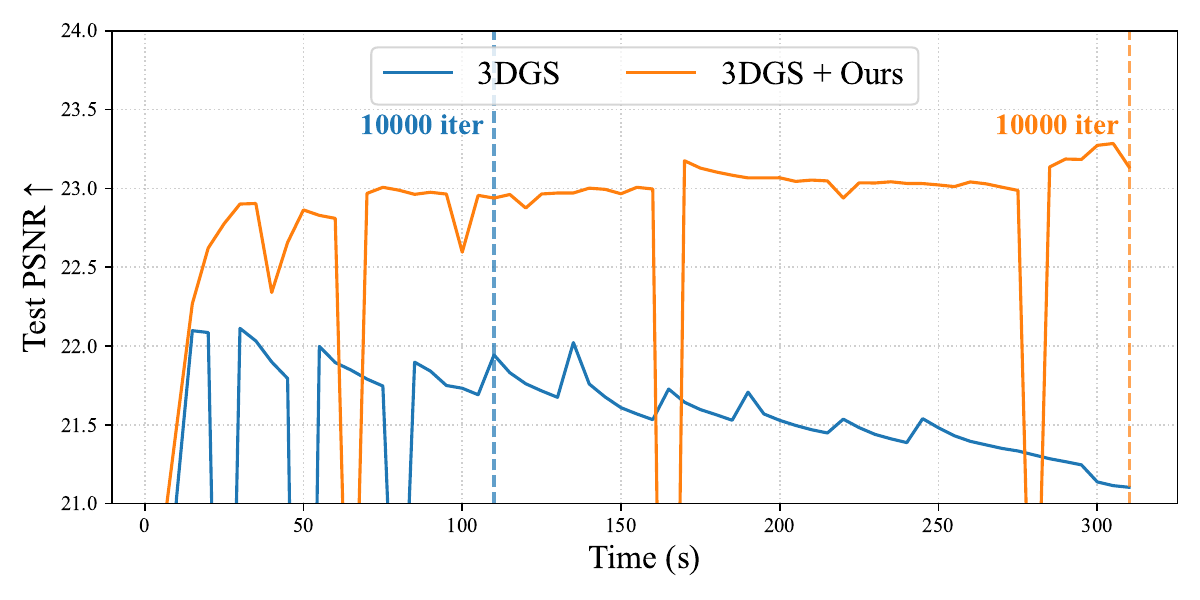}
    \caption{\textbf{Convergence speed}. We measure the test PSNR of each method for five seconds interval. Our method achieves higher test PSNR than the baseline within the same training time.}
    \label{fig:convergence}
\end{figure}


\section{Computational Cost}
\label{supple:cost}
The primary limitation of explicit flat minima optimization methods (e.g., SAM) is that they compute the loss gradient twice per iteration to estimate loss sharpness, doubling the training time per step. 
Despite inheriting this overhead, our approach achieves superior reconstruction quality compared to the baseline 3DGS within the same wall-clock time (\cref{fig:convergence}). 
Furthermore, the additional training time of our method leads to even more improvements, while 3DGS exhibits overfitting.

To further reduce the computational cost, we introduce a speedup strategy, denoted as Ours-L. 
Inspired by Zhou~\etal~\cite{zhou2025latesam}, we apply our method only during the final 12.5\% of total iterations. 
As shown in \cref{tab:late_phase}, Ours-L offers a trade-off between performance and training speed; while its AVGE improvement over the baseline (0.0095) is slightly less than that of the full method (0.0132), it effectively reduces the training time overhead from $2.77\times$ to $1.15\times$ compared to 3DGS.
Furthermore, we show that while applying it to late training iterations is most effective, early training iterations remain influential (\cref{fig:tradeoff}). We infer that this tendency arises from densification, which dynamically adjusts the number of model parameters.

\begin{table}[t!]
    \centering
    \caption{\textbf{Applying our method at late training phase}. 
    For efficient training, we apply our method during the later iterations, denoted as Ours-L. 
    We report metrics \scriptsize{(and their changes relative to 3DGS)} \normalsize{averaged over ten runs on an RTX A5000; standard deviations are omitted for brevity. 
    \textbf{Bold} indicates the best performance, and \underline{underline} indicates the second best.}}
    \begin{tabular}{lcccc}
    \toprule
    Method & AVGE~$\downarrow$ & Time (sec)~$\downarrow$ \\
    \midrule
    3DGS & .1111 & \textbf{85.7} \\
    \midrule
    3DGS + Ours & \textbf{.0979} \scriptsize{(-.0132)} & 237. \scriptsize{($2.77\times$)} \\
    3DGS + Ours-L & \underline{.1016} \scriptsize{(-.0095)} & \underline{98.2} \scriptsize{($1.15\times$)} \\
    \bottomrule
    \end{tabular}
    \label{tab:late_phase}
\end{table}

\begin{figure}[tb!]
    \centering
    \includegraphics[width=0.5\linewidth]{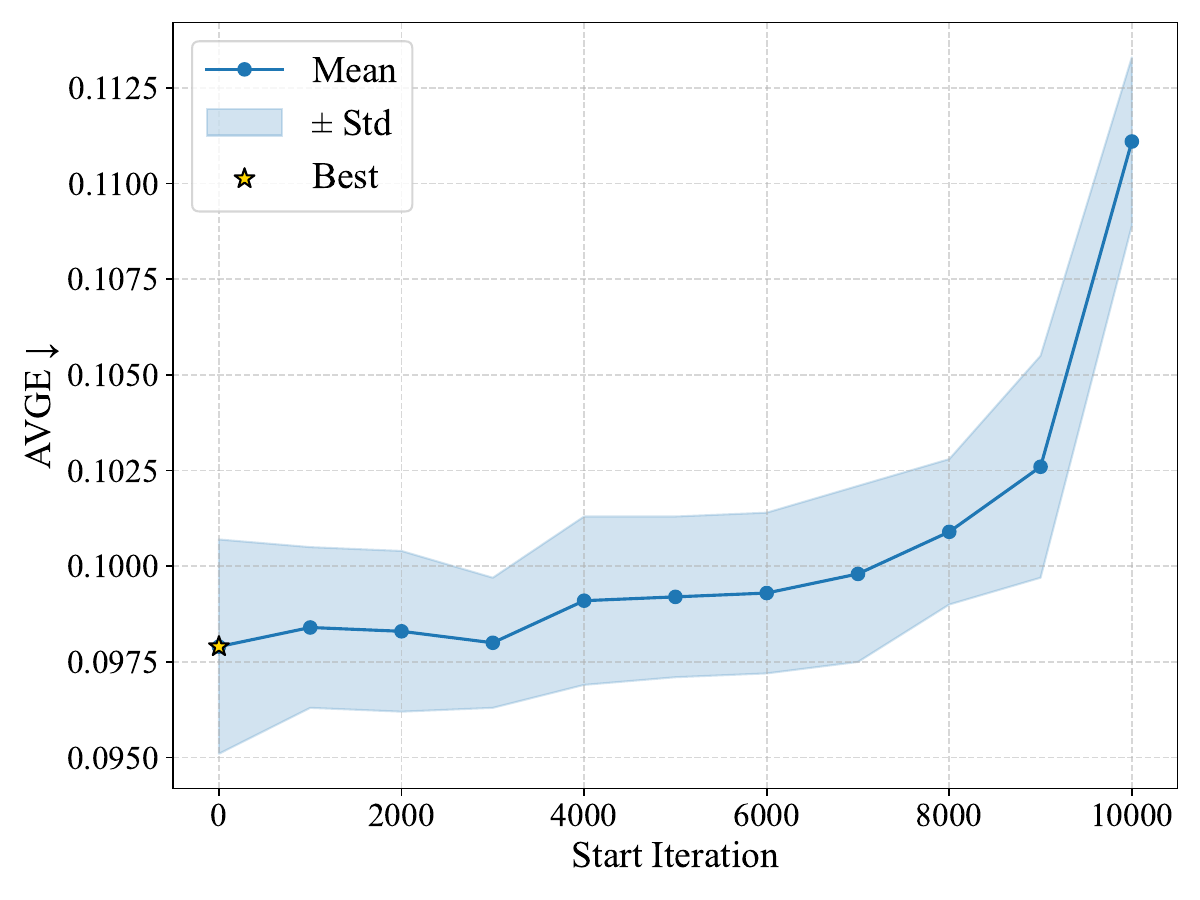}
    \caption{\textbf{Ablation study on the application phase}. We demonstrate the performance improvements obtained by varying the application duration in 10\% increments of the total training iterations.}
    \label{fig:tradeoff}
\end{figure}


\section{Analysis of Structure-Aware Flatness}
\label{supple:proof}
In this section, we analyze the loss landscape when assuming an ideal 3DGS solution that faithfully represents the underlying scene, providing a structure-aware characterization of the optimal neighborhood radius.
The key observation is that the admissible geometric perturbation of a Gaussian is correlated with the local image structure around its projection: Gaussians projected near high-detail structures admit smaller neighborhoods, whereas those projected onto low-detail structures admit larger ones.
We formalize this intuition using geometric tolerance and lift it to 3D under perspective projection.

In the following analysis, we consider the geometric parameter of Gaussian $\mathcal{G}_i$ to be its mean, i.e., $\boldsymbol{\theta}_i = \boldsymbol{\mu}_i$.
A similar argument applies to the scale parameter $s_i$, approximately due to its exponential activation, and we observe empirically that the proposed mechanism remains effective for the rotation parameter $\boldsymbol{q}_i$ as well.
Therefore, in practice, additional hyperparameters are required when applying the method to scale and rotation parameters.

\paragraph{Per-Gaussian local loss.}
For a training view $v$ and a Gaussian $\mathcal{G}_i$, let $L_i^v(\boldsymbol{\theta}_i)$ denote the $L_1$ rendering loss as a function of a geometric parameters $\boldsymbol{\theta}_i$ of $\mathcal{G}_i$, while all other Gaussian parameters are kept fixed.

\subsubsection{Geometric tolerance induced from local image structure.}
\label{sec:method:ddmap_analysis}

We define the set of high-detail pixels using an image-gradient threshold $\tau_I>0$:
$$
\mathcal{H}^v
=
\left\{
\mathbf{p}\ \middle|\ \|\nabla \tilde{I}^v(\mathbf{p})\|_2 \ge \tau_I
\right\}.
$$
Then the geometric tolerance map is given by
$\Gamma^v(\mathbf{p})
=
\min_{\mathbf{h}\in\mathcal{H}^v}\|\mathbf{p}-\mathbf{h}\|_2$. \newline
In practice, we use a bounded tolerance map
$$
\bar{\Gamma}^v(\mathbf{p})
=
\min\!\bigl(\Gamma^v(\mathbf{p}),\gamma_{\max}\bigr),
$$
where $\gamma_{\max}$ is a prescribed upper bound.
This induces the following image-space neighborhood:
$$
\mathcal{N}_{2D}^v(\mathbf{p})
=
\left\{
\hat{\mathbf{p}}
\;\middle|\;
\|\hat{\mathbf{p}}-\mathbf{p}\|_2 < \bar{\Gamma}^v(\mathbf{p})
\right\}.
$$
Since $\bar{\Gamma}^v(\mathbf{p}) \le \Gamma^v(\mathbf{p})$, every pixel in $\mathcal{N}_{2D}^v(\mathbf{p})$ lies outside the high-detail set.
Under a local smoothness approximation, this implies that $\tilde{I}^v$ varies slowly throughout $\mathcal{N}_{2D}^v(\mathbf{p})$, with local variation controlled by $\tau_I$.
Accordingly, for $\hat{\mathbf{p}}\in\mathcal{N}_{2D}^v(\mathbf{p})$, we obtain the local bound
\begin{equation}
\|\tilde{I}^v(\hat{\mathbf{p}})-\tilde{I}^v(\mathbf{p})\|_1
\lesssim
\tau_I \|\hat{\mathbf{p}}-\mathbf{p}\|_2
<
\tau_I \bar{\Gamma}^v(\mathbf{p})
\le
\tau_I \gamma_{\max}.
\label{eq:image_variation_bound}
\end{equation}
Therefore, when the pixel $\mathbf{p}$ moves within $\mathcal{N}_{2D}^v(\mathbf{p})$, the induced image variation is bounded by $\tau_I \gamma_{\max}$.
When this bound is small, the image remains nearly unchanged in this neighborhood.

\subsubsection{3D flat neighborhoods.}

For Gaussian $\mathcal{G}_i$ under view $v$, let $\boldsymbol{\mu}'^{v}_{i}$ be the projection of its mean. 
The corresponding tolerance is given by
$$
\gamma_i^v
=
\bar{\Gamma}^v(\boldsymbol{\mu}'^v_i).
$$
Let $d_i^v$ be the rendered depth of $\mathcal{G}_i$ and $f$ the focal length.
Under a narrow field-of-view assumption and a local linearization of the perspective projection, a small image-plane displacement approximately requires a 3D displacement scaled by $d_i^v/f$.
Based on this approximation, we define the neighborhood radius for the parameter $\boldsymbol{\theta}_i$ as
$$
\rho_{\boldsymbol{\theta}_i}
=
\gamma_i^v\frac{d_i^v}{f},
$$
and its 3D neighborhood is given by
$$
\mathcal{N}_{3D}^v(\mathcal{G}_i)
=
\left\{
\hat{\boldsymbol{\theta}}
\;\middle|\;
\|\hat{\boldsymbol{\theta}} - \boldsymbol{\theta}_i\|_2 \le \rho_{\boldsymbol{\theta}_i}
\right\}.
$$
Under this local approximation, perturbations in $\mathcal{N}_{3D}^v(\mathcal{G}_i)$ are mapped to image-space perturbations that remain within $\mathcal{N}_{2D}^v(\boldsymbol{\mu}'^v_i)$.

Assuming the ideal solution, i.e. $\text{Render}(\mathcal{G}, v)\approx \tilde{I}^v$, loss changes are expected to be dominated by the rendering variation.
Using this assumption, the image-space bound in \cref{eq:image_variation_bound} suggests the following local loss-variation bound:
$$
\bigl|
L_i^v(\hat{\boldsymbol{\theta}})
-
L_i^v(\boldsymbol{\theta}_i)
\bigr|
\lesssim
\tau_I \gamma_i^v
\le
\tau_I \gamma_{\max},
\qquad
\forall\, \hat{\boldsymbol{\theta}} \in \mathcal{N}_{3D}^v(\mathcal{G}_i).
$$
That is, for sufficiently small $\tau_I \gamma_{\max}$, $\mathcal{N}_{3D}^v(\mathcal{G}_i)$ provides a local estimate of the flat neighborhood (i.e., the neighborhood in which the loss remains nearly unchanged) around $\mathcal{G}_i$ under view $v$, with its size determined by the projected image structure.


\begin{table}[t!]
    \centering
    \caption{\textbf{Contribution of Gaussian attributes}. The Gaussian mean is the dominant contributor to the performance improvement.}
    \begin{tabular}{lccc}
    \toprule
    Method  &  PSNR~$\uparrow$     &  SSIM~$\uparrow$      &  LPIPS~$\downarrow$ \\
    \midrule
    3DGS + Ours & \textbf{20.783} \scriptsize{$\pm$ .300} & \textbf{.7197} \scriptsize{$\pm$ .0032} & \textbf{.1965} \scriptsize{$\pm$ .0034} \\
    \midrule
    without mean & 20.053 \scriptsize{$\pm$ .228} & .6903 \scriptsize{$\pm$ .0051} & .2102 \scriptsize{$\pm$ .0040} \\
    without rotation & 20.656 \scriptsize{$\pm$ .241} & .7175 \scriptsize{$\pm$ .0046} & .1976 \scriptsize{$\pm$ .0040} \\
    without scale & 20.641 \scriptsize{$\pm$ .235} & .7171 \scriptsize{$\pm$ .0049} & .1987 \scriptsize{$\pm$ .0035} \\
    without opacity & 20.628 \scriptsize{$\pm$ .218} & .7175 \scriptsize{$\pm$ .0044} & .1977 \scriptsize{$\pm$ .0033} \\
    without SH$_\text{DC}$ & 20.594 \scriptsize{$\pm$ .242} & .7152 \scriptsize{$\pm$ .0052} & .1972 \scriptsize{$\pm$ .0036} \\
    without SH$_\text{AC}$ & 20.599 \scriptsize{$\pm$ .205} & .7169 \scriptsize{$\pm$ .0041} & .1973 \scriptsize{$\pm$ .0034} \\
    \midrule
    only mean & 20.561 \scriptsize{$\pm$ .268} & .7147 \scriptsize{$\pm$ .0046} & .1975 \scriptsize{$\pm$ .0037} \\
    only rotation & 19.854 \scriptsize{$\pm$ .168} & .6828 \scriptsize{$\pm$ .0029} & .2110 \scriptsize{$\pm$ .0024} \\
    only scale & 20.036 \scriptsize{$\pm$ .222} & .6871 \scriptsize{$\pm$ .0048} & .2093 \scriptsize{$\pm$ .0037} \\
    only opacity & 19.888 \scriptsize{$\pm$ .188} & .6803 \scriptsize{$\pm$ .0045} & .2124 \scriptsize{$\pm$ .0033} \\
    only SH$_\text{DC}$ & 19.990 \scriptsize{$\pm$ .202} & .6867 \scriptsize{$\pm$ .0048} & .2115 \scriptsize{$\pm$ .0034} \\
    only SH$_\text{AC}$ & 19.914 \scriptsize{$\pm$ .186} & .6827 \scriptsize{$\pm$ .0042} & .2114 \scriptsize{$\pm$ .0033} \\
    \midrule
    3DGS & 19.810 \scriptsize{$\pm$ .339} & .6790 \scriptsize{$\pm$ .0078} & .2145 \scriptsize{$\pm$ .0065} \\
    \bottomrule
    \end{tabular}
    \label{tab:attribute}
\end{table}

\section{Effect of Individual Gaussian Attributes}
We analyze the contribution of each Gaussian attribute by applying our method either exclusively to a single attribute or to all attributes except one. While all attributes contribute positively to the final result, the Gaussian mean is the dominant contributor to the performance gain (\cref{tab:attribute}). This finding suggests that, instead of perturbing all attributes simultaneously, applying our method only to the Gaussian mean could serve as a simplified yet highly effective approach.


\begin{figure}[t]
    \centering
    \includegraphics[width=\textwidth]{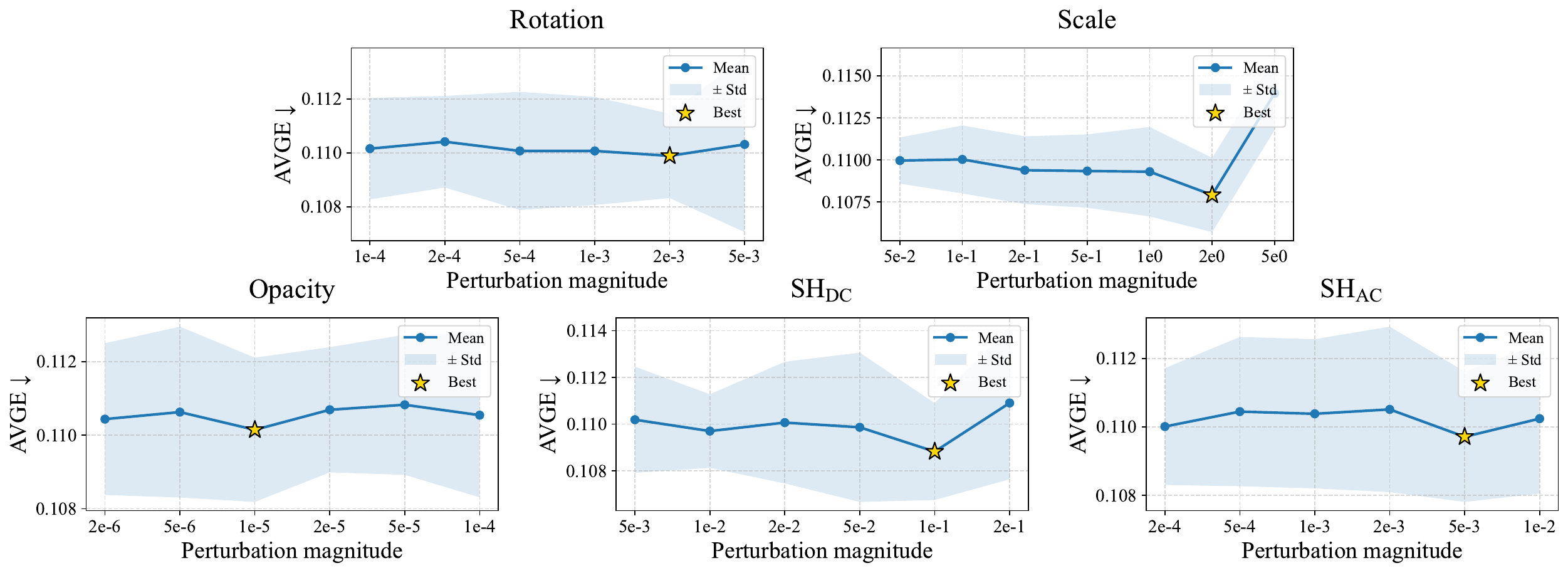}
    \caption{\textbf{Hyperparameter grid search, where each parameter is perturbed individually on the LLFF dataset}. Plots are means and standard deviations over ten runs. We mark a star at the best hyperparameter value.}
    \label{fig:grid_search}
\end{figure} 
\begin{table}[tb!]
    \centering
    \caption{\textbf{Sensitivity of maximum regularization weight $\bar{\gamma}^\text{max}$}. Performance differences across hyperparameter settings are minimal, suggesting limited benefit from hyperparameter tuning.}
    \begin{tabular}{lccc}
    \toprule
    $\bar{\gamma}^\text{max}$  &  PSNR~$\uparrow$     &  SSIM~$\uparrow$      &  LPIPS~$\downarrow$ \\
    \midrule
    0.95 & \textbf{20.783} \scriptsize{$\pm$ .300} & \textbf{.7197} \scriptsize{$\pm$ .0032} & \textbf{.1965} \scriptsize{$\pm$ .0034} \\
    0.80 & 20.780 \scriptsize{$\pm$ .223} & .7195 \scriptsize{$\pm$ .0049} & .1971 \scriptsize{$\pm$ .0036} \\
    0.65 & 20.721 \scriptsize{$\pm$ .264} & .7187 \scriptsize{$\pm$ .0044} & .1968 \scriptsize{$\pm$ .0038} \\
    0.50 & 20.730 \scriptsize{$\pm$ .171} & .7190 \scriptsize{$\pm$ .0040} & .1968 \scriptsize{$\pm$ .0032} \\
    \bottomrule
    \end{tabular}
    \label{tab:grid_gamma}
\end{table}

\section{Choice of Hyperparameters} 
\label{supple:grid_search}
Determining the optimal neighborhood radius $\rho$, i.e., perturbation magnitude, is a challenge in explicit flat minima optimization. 
Notably, our method replaces the hyperparameter for the Gaussian mean with its corresponding geometric tolerance (\cref{supple:proof}), fixing $\rho_\mu$ to 1. 
For other Gaussian attributes, the method scales the $\rho$; $\rho$ itself remains a core hyperparameter.

To determine the optimal $\rho$, we initially perform a grid search by perturbing each parameter independently (\cref{fig:grid_search}). However, because the parameters are interdependent and mutually influence one another, the optimal $\rho$ for simultaneous perturbation is generally smaller than the individually derived values. Therefore, we perform a subsequent grid search among these smaller candidate values to determine the final $\rho$.

Regarding the maximum regularization weight (\cref{sec:method:fasr:sharp}), we selected the $\bar{\gamma}^\text{max}$ with the best performance during grid search (\cref{tab:grid_gamma}); as the model performance remains robust across a broad spectrum of values, extensive tuning of this weight is unnecessary in practice.

\begin{figure}[t]
    \centering
    \includegraphics[width=\textwidth]{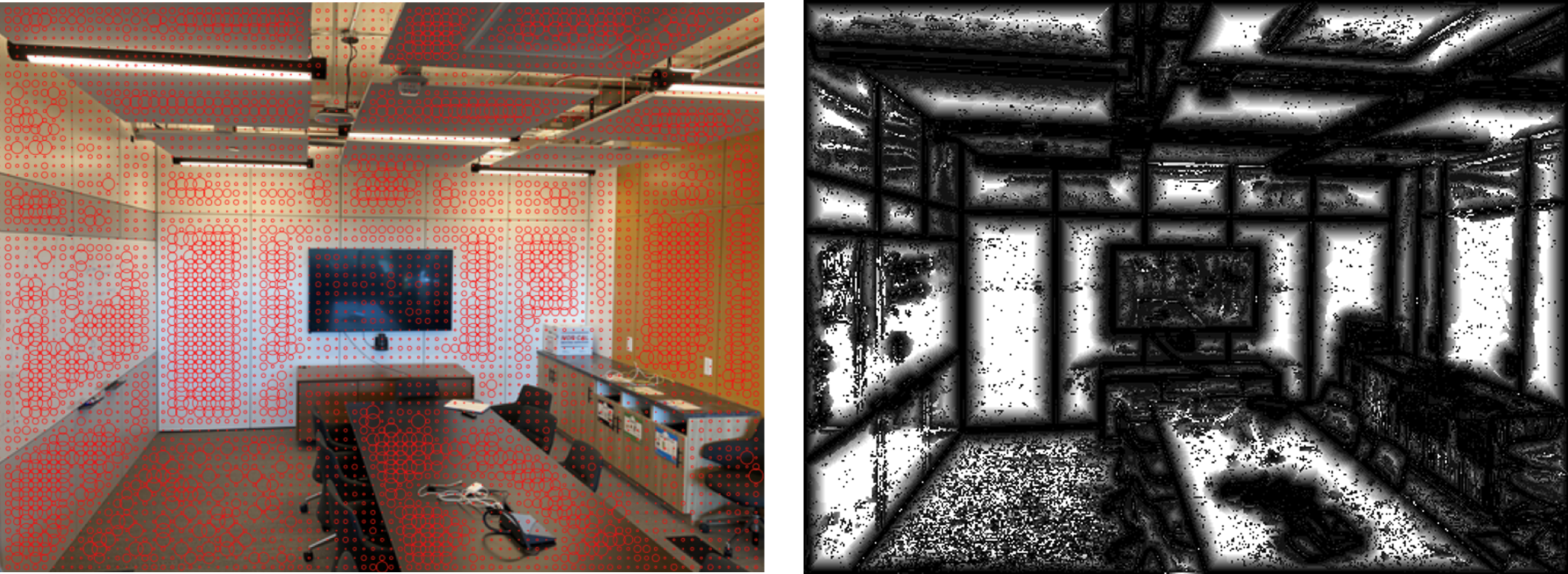}
    \caption{\textbf{Visualization of the geometric tolerance map}. Left: tolerance values overlaid on the image using circles whose radii correspond to the pixel-wise tolerance.
    Right: grayscale visualization of the tolerance map, where brighter intensities indicate larger tolerance values.}
    \label{fig:tolerance_map}
\end{figure}

\section{Design Choice of Geometric Tolerance Map}
\label{supple:log}

In practice, we implement the geometric tolerance map using a multi-scale analysis inspired by blob detection with the Laplacian of Gaussian (LoG)~\cite{lindeberg2013log}. 
Although this implementation is not identical to the idealized definition in the main text, it exhibits similar behavior in practice by assigning smaller tolerance values near high-detail structures and larger values in low-detail structures (see the visualization in \cref{fig:tolerance_map}).

Given a grayscale image $\tilde{I}_\text{gray}$, we compute the absolute LoG response at a set of candidate scales $\sigma_1 < \sigma_2 < \dots < \sigma_K$.
The largest scale is used as the maximum tolerance, i.e., $\gamma_{\max}=\sigma_K$.
At each pixel $\mathbf{p}$, we examine the response curve across scales and detect the first significant increase:
$$
k
=
\argmin_{j=1,\dots,K-1}
\left\{
\left|
\mathrm{LoG}_{\sigma_{j+1}}(\tilde{I}_\text{gray})(\mathbf{p})
\right|
-
\left|
\mathrm{LoG}_{\sigma_j}(\tilde{I}_\text{gray})(\mathbf{p})
\right|
>
\tau_{\mathrm{LoG}}
\right\},
$$
where $\tau_{\mathrm{LoG}}$ is a threshold.
We then assign the geometric tolerance at $\mathbf{p}$ as the smallest scale preceding this rise,
$$
\Gamma^v(\mathbf{p})=\sigma_{k-1}.
$$
If no such increase is observed, the tolerance is set to $\Gamma^v(\mathbf{p})=\sigma_K$.
Applying this procedure to all pixels yields the geometric tolerance map
$\boldsymbol{\Gamma}^v$ for view $v$.

We also experimented with alternative implementations, including the determinant of Hessian (DoH) and distance transform (DT).
The quantitative comparison is reported in \cref{tab:operator}.
Among these choices, the LoG-based implementation performed best overall, and we therefore use it in all experiments.

\begin{table}[tb!]
    \centering
    \caption{\textbf{Ablation study on the operator of geometric tolerance map generation}. We select the best-performing LoG-based implementation.}
    \begin{tabular}{lccc}
    \toprule
    Operator  &  PSNR~$\uparrow$     &  SSIM~$\uparrow$      &  LPIPS~$\downarrow$ \\
    \midrule
    LoG & \textbf{20.783} \scriptsize{$\pm$ .300} & \textbf{.7197} \scriptsize{$\pm$ .0032} & \textbf{.1965} \scriptsize{$\pm$ .0034} \\
    DoH & 20.646 \scriptsize{$\pm$ .336} & .7143 \scriptsize{$\pm$ .0041} & .1988 \scriptsize{$\pm$ .0046} \\
    DT & 20.697 \scriptsize{$\pm$ .208} & .7140 \scriptsize{$\pm$ .0030} & .1980 \scriptsize{$\pm$ .0028} \\
    \bottomrule
    \end{tabular}
    \label{tab:operator}
\end{table}

\section{Implementation detail}
For all methods except 3DGS, we use the official repository. Since the official 3DGS does not target sparse novel view synthesis, we refactor it by referring to CoR-GS and DropGaussian. NexusGS does not provide optical flow for MipNeRF-360; we compute the flow using FlowFormer++~\cite{shi2023flowformerpp}. 
When the baseline utilizes regularization, we backpropagate it after computing the gradient of our proposed method, and for the deformation process, we compute the gradient on the deformed Gaussians and propagate it back. 
We update the residual map of Yun \etal~\cite{yun2025or2} during the ascent step and the Gaussians during the descent step. 
For each dataset, we set the same values of $\rho_\theta$ and $\gamma$ for all baselines except NexusGS, where we set them to $0.1\times$ the values used for other baselines due to its use of denser per-pixel Gaussians.

\begin{figure}[tb!]
    \centering
    \includegraphics[width=\linewidth]{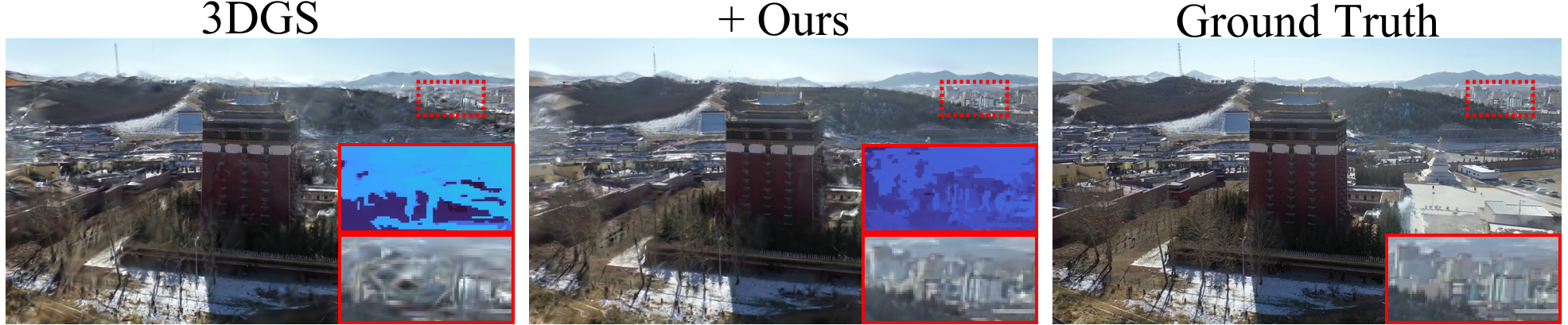}
    \caption{\textbf{Qualitative results on scene \texttt{15} from the OMMO dataset.} Our method is effective in a wide depth range of regions.}
    \label{fig:ommo}
\end{figure}
\begin{table}[t!]
    \centering
    \caption{\textbf{Quantitative evaluation on scene \texttt{15} of the OMMO dataset}. Our method is effective in large scene scenario with severe scale ambiguity.}
    \begin{tabular}{lccc}
    \toprule
         Method & PSNR $\uparrow$ & SSIM $\uparrow$ & LPIPS $\downarrow$  \\
         \midrule
         3DGS & 19.134 \scriptsize{$\pm$ .113} & .6284 \scriptsize{$\pm$ .0050} & .2965 \scriptsize{$\pm$ .0048} \\
         + Ours & \textbf{19.928} \scriptsize{$\pm$ .174} & \textbf{.6580} \scriptsize{$\pm$ .0046} & \textbf{.2696} \scriptsize{$\pm$ .0044} \\
         \bottomrule
    \end{tabular}
    \label{tab:ommo}
\end{table}

\section{More results}
\subsection{Large Scene Scenario}
We further evaluate our method on scenes with a significantly larger scale than the unbounded scenes in Mip-NeRF 360 (e.g., \texttt{bicycle} scene). Specifically, we conducted an experiment on \texttt{15} scene of OMMO dataset~\cite{lu2023ommo}. As shown in \cref{fig:ommo} and \cref{tab:ommo}, our method is effective in this large scale ambiguity environment.

\begin{table}[t]
    \centering
    \caption{\textbf{Quantitative results under various input view settings on the LLFF dataset.} Our method yields greater improvement in highly sparse configurations.}
        \begin{tabular}{lccccccccc}
        \toprule
         & \multicolumn{3}{c}{2-view} & \multicolumn{3}{c}{6-view} & \multicolumn{3}{c}{9-view} \\ \cmidrule(lr){2-4}\cmidrule(lr){5-7}\cmidrule(lr){8-10}
         Method & PSNR $\uparrow$ & SSIM $\uparrow$ & LPIPS $\downarrow$ & PSNR $\uparrow$ & SSIM $\uparrow$ & LPIPS $\downarrow$ & PSNR $\uparrow$ & SSIM $\uparrow$ & LPIPS $\downarrow$ \\ \midrule
         3DGS & 18.46 & .6076 & .2581 & 24.05 & .8112 & .1259 & 25.54 & .8594 & .1045 \\
         +Ours & \textbf{19.40} & \textbf{.6667} & \textbf{.2275} & \textbf{24.54} & \textbf{.8321} & \textbf{.1112} & \textbf{25.99} & \textbf{.8653} & \textbf{.0924} \\ 
         \bottomrule
        \end{tabular}
\label{tab:more_view}
\end{table}
\begin{figure}[tb!]
    \centering
    \includegraphics[width=\linewidth]{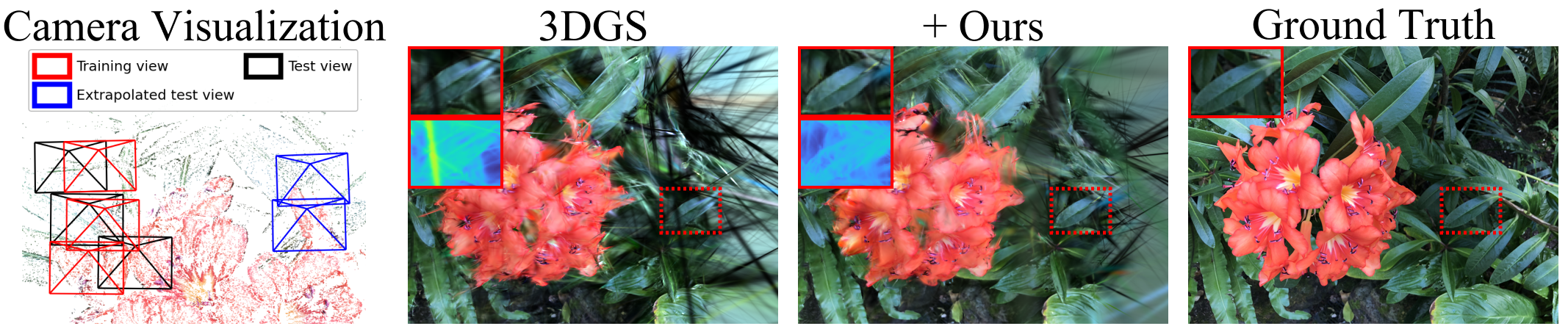}
    \caption{\textbf{Qualitative results on challenging extrapolated views}. Our method remains effective in these views, though some artifacts remain in regions without training view overlap.}
    \label{fig:extrap}
\end{figure}

\begin{table}[t]
    \centering
    \caption{\textbf{Quantitative comparison of post-optimization methods for feed-forward 3D reconstruction on the 3-view DTU dataset}. Applying our method on the post-optimization step improves reconstruction quality.}
    \begin{tabular}{ccccc}
    \toprule
    Baseline & Post-Optimization  & PSNR $\uparrow$ & SSIM $\uparrow$ & LPIPS $\downarrow$  \\ \midrule
    \multirow{3}{*}{AnySplat}    & \xmark & 12.629 & .5191 & .3573 \\
     & 3DGS & 13.239 & .5232 & .3597 \\
     & 3DGS + Ours & \textbf{15.530} & \textbf{.5576} & \textbf{.3538} \\
    \bottomrule
    \end{tabular}
    \label{tab:feedforward}
\end{table}

\subsection{Test Views on Extreme Viewpoints}
Beyond analyzing results based on input sparsity, we further evaluate challenging test views. As shown in the first column of \cref{fig:extrap}, \texttt{flower} scene of LLFF dataset contains extrapolated views. As demonstrated in the second and third columns of \cref{fig:extrap}, our method remains effective even under these challenging viewpoints. Nevertheless, we observe a failure case in which artifacts persist in regions that are unobserved by the training views. Improving rendering quality in these non-overlapping regions necessitates additional priors, which we leave as future work.

\subsection{Various Number of Input Views}
To further support the finding in \cref{sec:exp:analysis:covis} that our proposed method is more effective in sparser settings, we conducted additional experiments with a wider range of input view configurations. 
In addition to the 3-view setting on the LLFF dataset in the main experiments, we evaluated our method on a more challenging 2-view setting, as well as on 6-view and 9-view settings commonly used in the prior works. As shown in \cref{tab:more_view}, our method yields greater performance advantages in sparser settings.

\subsection{Leveraging Feedforward Priors}
Recent advances in 3D reconstruction are shifting the paradigm from per-scene optimization toward feed-forward inference, even under sparse-view constraints. 
Since feed-forward models often employ post-optimization to improve their results, we show that our method remains effective in this setup. 
We evaluate our approach by applying it in the post-optimization step of AnySplat~\cite{jiang2025anysplat} on the DTU dataset~\cite{jensen2014dtu}. 
As shown in \cref{tab:feedforward}, our method improves reconstruction quality, demonstrating its ability when leveraging feed-forward priors.

\subsection{Per-Scene Results}
We report the quantitative results of each scene in \cref{tab:per-scene}. Our method method outperforms the baseline in most cases.

\begin{table*}[t!]
    \centering
    \caption{\textbf{Per-scene quantitative results on LLFF and MipNeRF-360 dataset}. We report AVGE (lower is better) of each scene.}
    \resizebox{\textwidth}{!}{%
    \begin{tabular}{lcccccccc}
    \toprule
    & \multicolumn{8}{c}{LLFF (3 views)} \\ \cmidrule(rl){2-9}
    Method & \texttt{fern} & \texttt{flower} & \texttt{fortress} & \texttt{horns} & \texttt{leaves} & \texttt{orchids} & \texttt{room} & \texttt{trex} \\
    \cmidrule{1-9}
    3DGS & .0950 \scriptsize{$\pm$.01790} & .1147 \scriptsize{$\pm$.00180} & .0885 \scriptsize{$\pm$.00390} & .1276 \scriptsize{$\pm$.00240} & .1362 \scriptsize{$\pm$.00170} & .1769 \scriptsize{$\pm$.00190} & .0753 \scriptsize{$\pm$.00160} & .0749 \scriptsize{$\pm$.00210} \\
    + Ours & \textbf{.0765} \scriptsize{$\pm$.00080} & \textbf{.1066} \scriptsize{$\pm$.00140} & \textbf{.0781} \scriptsize{$\pm$.00530} & \textbf{.1044} \scriptsize{$\pm$.00280} & \textbf{.1170} \scriptsize{$\pm$.00130} & \textbf{.1605} \scriptsize{$\pm$.00190} & \textbf{.0686} \scriptsize{$\pm$.00190} & \textbf{.0712} \scriptsize{$\pm$.00710} \\
    \cmidrule{1-9}
    CoR-GS & .0806 \scriptsize{$\pm$.00080} & .1088 \scriptsize{$\pm$.00290} & .0754 \scriptsize{$\pm$.00160} & .1205 \scriptsize{$\pm$.00120} & .1427 \scriptsize{$\pm$.00310} & .1735 \scriptsize{$\pm$.00200} & .0723 \scriptsize{$\pm$.00130} & .0699 \scriptsize{$\pm$.00440} \\
    + Ours & \textbf{.0724} \scriptsize{$\pm$.00090} & \textbf{.1014} \scriptsize{$\pm$.00220} & \textbf{.0683} \scriptsize{$\pm$.00190} & \textbf{.1075} \scriptsize{$\pm$.00200} & \textbf{.1326} \scriptsize{$\pm$.00240} & \textbf{.1603} \scriptsize{$\pm$.00100} & \textbf{.0695} \scriptsize{$\pm$.00100} & \textbf{.0614} \scriptsize{$\pm$.00340} \\
    \cmidrule{1-9}
    DropGaussian & .0791 \scriptsize{$\pm$.00120} & .1056 \scriptsize{$\pm$.00210} & .0796 \scriptsize{$\pm$.00420} & .1186 \scriptsize{$\pm$.00270} & .1284 \scriptsize{$\pm$.00190} & .1650 \scriptsize{$\pm$.00160} & .0715 \scriptsize{$\pm$.00150} & .0687 \scriptsize{$\pm$.00190} \\
    + Ours & \textbf{.0717} \scriptsize{$\pm$.00100} & \textbf{.1048} \scriptsize{$\pm$.00170} & \textbf{.0784} \scriptsize{$\pm$.00340} & \textbf{.1101} \scriptsize{$\pm$.00340} & \textbf{.1176} \scriptsize{$\pm$.00140} & \textbf{.1536} \scriptsize{$\pm$.00170} & \textbf{.0701} \scriptsize{$\pm$.00200} & \textbf{.0657} \scriptsize{$\pm$.00260} \\
    \cmidrule{1-9}
    NexusGS & .0893 \scriptsize{$\pm$.00030} & \textbf{.0981} \scriptsize{$\pm$.00020} & .0498 \scriptsize{$\pm$.00030} & .0909 \scriptsize{$\pm$.00050} & .1072 \scriptsize{$\pm$.00030} & .1383 \scriptsize{$\pm$.00060} & .0755 \scriptsize{$\pm$.00060} & .0810 \scriptsize{$\pm$.00050} \\
    + Ours & \textbf{.0848} \scriptsize{$\pm$.00050} & .0989 \scriptsize{$\pm$.00060} & \textbf{.0475} \scriptsize{$\pm$.00090} & \textbf{.0897} \scriptsize{$\pm$.00070} & \textbf{.0993} \scriptsize{$\pm$.00060} & \textbf{.1354} \scriptsize{$\pm$.00040} & \textbf{.0696} \scriptsize{$\pm$.00030} & \textbf{.0741} \scriptsize{$\pm$.00060} \\
    \cmidrule{1-9}
    SE-GS & .0720 \scriptsize{$\pm$.00040} & .1052 \scriptsize{$\pm$.00180} & .0662 \scriptsize{$\pm$.00130} & .1104 \scriptsize{$\pm$.00180} & .1392 \scriptsize{$\pm$.00230} & .1707 \scriptsize{$\pm$.01440} & .0635 \scriptsize{$\pm$.00220} & .0596 \scriptsize{$\pm$.00180} \\
    + Ours & \textbf{.0675} \scriptsize{$\pm$.00060} & \textbf{.0992} \scriptsize{$\pm$.00160} & \textbf{.0661} \scriptsize{$\pm$.00820} & \textbf{.1028} \scriptsize{$\pm$.00160} & \textbf{.1310} \scriptsize{$\pm$.00160} & \textbf{.1546} \scriptsize{$\pm$.00170} & \textbf{.0629} \scriptsize{$\pm$.00120} & \textbf{.0564} \scriptsize{$\pm$.00080} \\
    \cmidrule{1-9}
    & \multicolumn{7}{c}{MipNeRF-360 (12 views)} & \multicolumn{1}{c}{} \\ \cmidrule(rl){2-8}
    Method & \texttt{bicycle} & \texttt{bonsai} & \texttt{counter} & \texttt{garden} & \texttt{kitchen} & \texttt{room} & \texttt{stump} &  \\
    \cmidrule{1-8}
    3DGS & .1763 \scriptsize{$\pm$.00400} & .1408 \scriptsize{$\pm$.00400} & .1464 \scriptsize{$\pm$.00250} & .1302 \scriptsize{$\pm$.00190} & .1052 \scriptsize{$\pm$.00160} & .1078 \scriptsize{$\pm$.00110} & .2347 \scriptsize{$\pm$.00280} &  \\
    + Ours & \textbf{.1714} \scriptsize{$\pm$.00120} & \textbf{.1285} \scriptsize{$\pm$.00240} & \textbf{.1350} \scriptsize{$\pm$.00110} & \textbf{.1240} \scriptsize{$\pm$.00130} & \textbf{.1017} \scriptsize{$\pm$.00160} & \textbf{.1021} \scriptsize{$\pm$.00460} & \textbf{.2253} \scriptsize{$\pm$.00640} &  \\
    \cmidrule{1-8}
    CoR-GS & .1734 \scriptsize{$\pm$.00340} & .1262 \scriptsize{$\pm$.00250} & .1310 \scriptsize{$\pm$.00120} & .1297 \scriptsize{$\pm$.00370} & .1054 \scriptsize{$\pm$.00360} & \textbf{.0931} \scriptsize{$\pm$.00230} & .2214 \scriptsize{$\pm$.00770} &  \\
    + Ours & \textbf{.1733} \scriptsize{$\pm$.00170} & \textbf{.1131} \scriptsize{$\pm$.00230} & \textbf{.1229} \scriptsize{$\pm$.00060} & \textbf{.1287} \scriptsize{$\pm$.00290} & \textbf{.1044} \scriptsize{$\pm$.00320} & .0993 \scriptsize{$\pm$.00440} & \textbf{.2073} \scriptsize{$\pm$.00950} &  \\
    \cmidrule{1-8}
    DropGaussian & .1646 \scriptsize{$\pm$.00110} & .1301 \scriptsize{$\pm$.00210} & .1356 \scriptsize{$\pm$.00140} & .1243 \scriptsize{$\pm$.00110} & \textbf{.0979} \scriptsize{$\pm$.00270} & .1026 \scriptsize{$\pm$.00500} & .2187 \scriptsize{$\pm$.00410} &  \\
    + Ours & \textbf{.1613} \scriptsize{$\pm$.00310} & \textbf{.1286} \scriptsize{$\pm$.00070} & \textbf{.1309} \scriptsize{$\pm$.00100} & \textbf{.1218} \scriptsize{$\pm$.00200} & .1027 \scriptsize{$\pm$.00210} & \textbf{.1003} \scriptsize{$\pm$.00440} & \textbf{.2171} \scriptsize{$\pm$.01180} &  \\
    \cmidrule{1-8}
    NexusGS & .1818 \scriptsize{$\pm$.00100} & .1308 \scriptsize{$\pm$.00190} & .1563 \scriptsize{$\pm$.00130} & .1249 \scriptsize{$\pm$.00030} & .0941 \scriptsize{$\pm$.00070} & .1506 \scriptsize{$\pm$.00380} & .2352 \scriptsize{$\pm$.00240} &  \\
    + Ours & \textbf{.1739} \scriptsize{$\pm$.00100} & \textbf{.1278} \scriptsize{$\pm$.00060} & \textbf{.1531} \scriptsize{$\pm$.00160} & \textbf{.1243} \scriptsize{$\pm$.00040} & \textbf{.0921} \scriptsize{$\pm$.00090} & \textbf{.1421} \scriptsize{$\pm$.00450} & \textbf{.2299} \scriptsize{$\pm$.00210} &  \\
    \cmidrule{1-8}
    SE-GS & .1624 \scriptsize{$\pm$.00350} & .1179 \scriptsize{$\pm$.00300} & .1251 \scriptsize{$\pm$.00370} & .1207 \scriptsize{$\pm$.00160} & .1107 \scriptsize{$\pm$.00370} & \textbf{.0875} \scriptsize{$\pm$.00380} & .2014 \scriptsize{$\pm$.00830} &  \\
    + Ours & \textbf{.1599} \scriptsize{$\pm$.00690} & \textbf{.1105} \scriptsize{$\pm$.00300} & \textbf{.1217} \scriptsize{$\pm$.00170} & \textbf{.1179} \scriptsize{$\pm$.00250} & \textbf{.1092} \scriptsize{$\pm$.00350} & .0973 \scriptsize{$\pm$.00090} & \textbf{.1941} \scriptsize{$\pm$.00480} &  \\
    \cmidrule[\heavyrulewidth]{1-8}
    \end{tabular}}
    \label{tab:per-scene}
\end{table*}

\section{Discussion}
While effective, our work has limitations.

First, manual hyperparameter tuning remains, except for the Gaussian mean. Specifically, for Gaussian scale and rotation, the base perturbation radius $\rho$ still requires tuning, despite adaptive adjustment. Additionally, handling non-geometric attributes (opacity and spherical harmonics) will require new strategies, which will be necessary for challenging conditions such as highly specular or transparent environments.
However, applying our structure-aware sharpness regularization exclusively to the Gaussian mean already yields highly competitive performance, suggesting that the additional performance gains from tuning the remaining Gaussian attributes would likely be marginal.

Lastly, the application of our strategy to distinct emerging works, such as feed-forward reconstruction models or employing generative priors, remains underexplored. However, given the consistent improvements demonstrated across a wide range of 3DGS-based frameworks, NeRFs, and dynamic scenes, we believe that our method will similarly benefit these approaches, leaving this as a promising direction for future work.
\fi
\fi
\end{document}